\documentclass{bmvc2k}
\usepackage{microtype}
\usepackage{graphicx}
\usepackage{subfigure}
\usepackage{booktabs} 
\usepackage{hyperref}

\usepackage{algorithm}
\usepackage{algorithmic}
\usepackage{amsmath}
\usepackage{amssymb}
\usepackage{mathtools}
\usepackage{amsthm}
\usepackage{physics}
\usepackage{caption}
\usepackage{rotating}
\usepackage{multirow}
\usepackage{pifont}
\usepackage[capitalize,noabbrev]{cleveref}
\usepackage{graphicx}

\theoremstyle{plain}

\theoremstyle{definition}

\theoremstyle{remark}

\usepackage{xcolor}
\usepackage{bm}

\newcommand{\epsilonb}{\boldsymbol{\epsilon}}

\newcommand{\comment}[1]{}
\title{LDEdit: Towards Generalized Text Guided\\ Image Manipulation via Latent Diffusion Models}
\addauthor{Paramanand Chandramouli}{paramanand.chandramouli[at]uni-siegen.de}{1}
\addauthor{Kanchana Vaishnavi Gandikota}{kanchana.gandikota[at]uni-siegen.de}{1}
\addinstitution{
 Department of Computer Science\\
 University of Siegen\\
 Germany
}
\runninghead{Chandramouli, Gandikota}{LDEdit: Towards Generalized}
\def\eg{\emph{e.g}\bmvaOneDot}

\begin{document}

\maketitle

\begin{abstract}
Research in vision-language models has seen rapid developments off-late, enabling natural language-based interfaces for image generation and manipulation. Many existing text guided manipulation techniques are restricted to specific classes of images, and often require fine-tuning to transfer to a different style or domain.  Nevertheless, generic image manipulation using a single model with flexible text inputs is highly desirable. Recent work  addresses this task by guiding  generative models trained on the generic image datasets using pretrained vision-language encoders.  While promising, this approach requires expensive optimization for each input. In this work, we propose an optimization-free method for the task of generic image manipulation from text prompts. Our approach  exploits recent Latent Diffusion Models~(LDM) for text to image generation to achieve zero-shot text guided manipulation. We employ a deterministic forward diffusion in a lower dimensional latent space, and the desired manipulation is achieved by simply providing the target text to condition the reverse diffusion process. We refer to our approach as LDEdit. We demonstrate the applicability of our method on  semantic image manipulation and artistic style transfer. Our method can accomplish image manipulation on diverse domains and enables editing multiple attributes in a straightforward fashion. Extensive experiments demonstrate the benefit of our approach over competing baselines.
\end{abstract}
\section{Introduction}\label{sec:intro}
Using natural language descriptions is an intuitive and  easy way for humans to communicate visual concepts. 
Hence, a tool which can automatically manipulate images using textual descriptions can greatly ease editing. This requires a careful control to modify only the relevant semantic attributes and styles while preserving the desired content representations. However, accomplishing this is highly challenging, especially when manipulating  open-domain images using arbitrary text prompts. As a result, many existing works allow manipulations which are restricted to a specific image classes~\cite{patashnik2021styleclip,xia2021tedigan,gal2021stylenada,kim2021diffusionclip} or a specific manipulation task~\cite{avrahami2021blended,nichol2021glide,kwon2021clipstyler}. 
Further, some of these methods require fine-tuned models \cite{kim2021diffusionclip,gal2021stylenada} for specific text prompts, further limiting their utility for flexible open domain image manipulation. In contrast to these techniques, the works \cite{liu2020open,crowson2022vqgan} handle general image manipulation from text prompts. While \cite{liu2020open} focuses on semantically simple transformations, \cite{crowson2022vqgan} allows more general text-to-image generation as well as manipulation using an expensive latent space optimization.\par
\begin{figure}[t]
\begin{minipage}[c]{0.49\textwidth}
\centering
\scriptsize{Input photo\enskip red lipcolor\quad+rose on hat\quad cartoon\quad+rose on hat}\\
\includegraphics[width=0.19\linewidth]{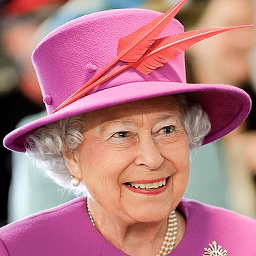}
\includegraphics[width=0.19\linewidth]{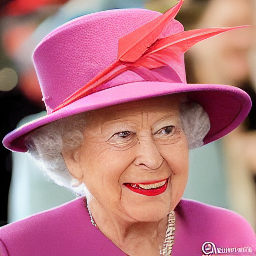}
\includegraphics[width=0.19\linewidth]{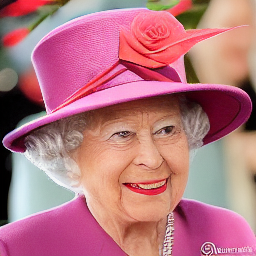}
\includegraphics[width=0.19\linewidth]{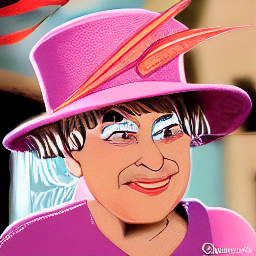}
\includegraphics[width=0.19\linewidth]{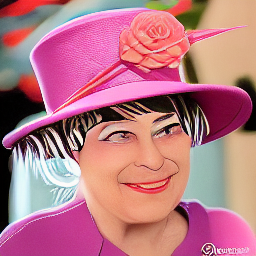}\\
\scriptsize{Input portrait~wrinkled skin\enskip+smiling\quad pixar+glasses\enskip van Gogh}\\
\includegraphics[width=0.19\linewidth]{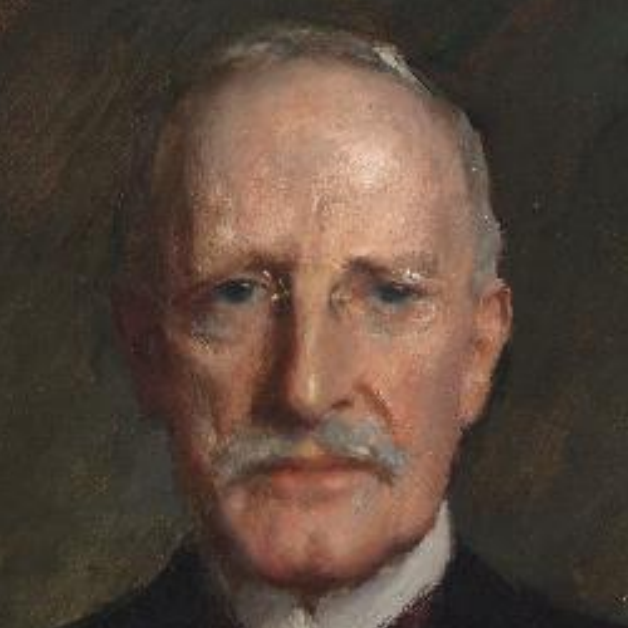}
\includegraphics[width=0.19\linewidth]{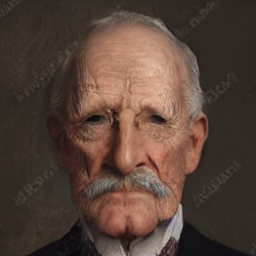}
\includegraphics[width=0.19\linewidth]{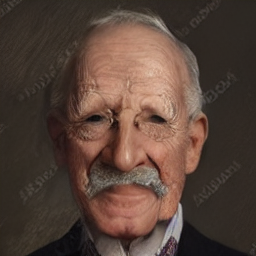}
\includegraphics[width=0.19\linewidth]{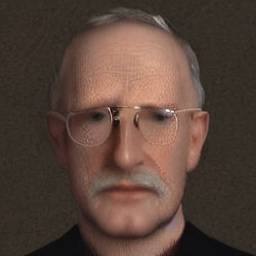}
\includegraphics[width=0.19\linewidth]{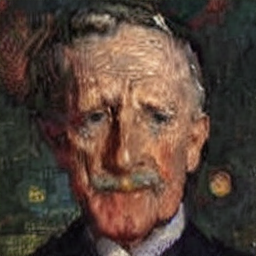}\\
\footnotesize{a) Editing local semantic attributes and global style}\vspace{2pt}\\
\scriptsize{Input stroke\quad old man\qquad woman\qquad pixar woman\enskip van Gogh}\\
\includegraphics[width=0.19\linewidth]{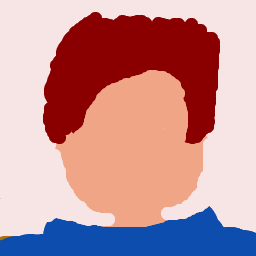}
\includegraphics[width=0.19\linewidth]{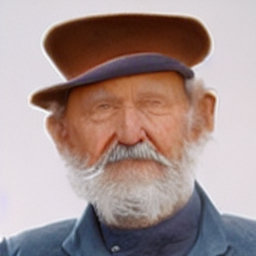}
\includegraphics[width=0.19\linewidth]{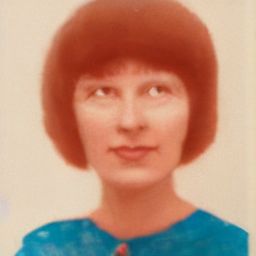}
\includegraphics[width=0.19\linewidth]{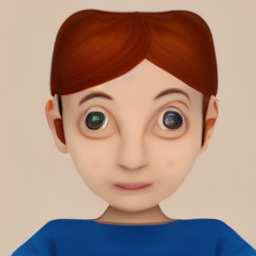}
\includegraphics[width=0.19\linewidth]{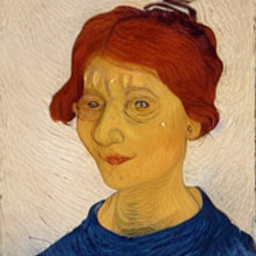}
\footnotesize{c) Stroke to image translation from text}\vspace{2pt}\\
\end{minipage}
\begin{minipage}[c]{0.49\textwidth}
\centering
\scriptsize{\quad Input\qquad red brick\quad wooden\quad Asian temple\quad+snow}\\
\includegraphics[width=0.19\linewidth]{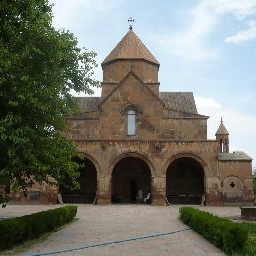}
\includegraphics[width=0.19\linewidth]{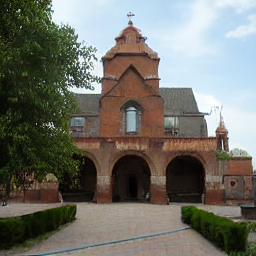}
\includegraphics[width=0.19\linewidth]{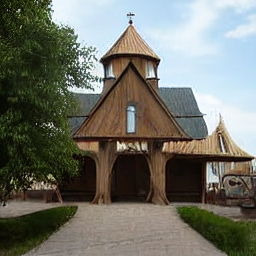}
\includegraphics[width=0.19\linewidth]{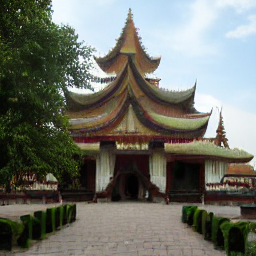}
\includegraphics[width=0.19\linewidth]{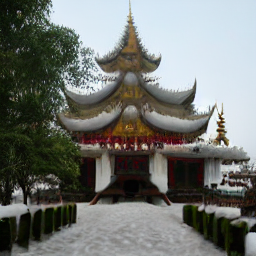}\\
\scriptsize{\enskip Input\qquad\enskip an eagle\quad a kingfisher\quad crow+tree\quad crow+sketch}\\
\includegraphics[width=0.19\linewidth]{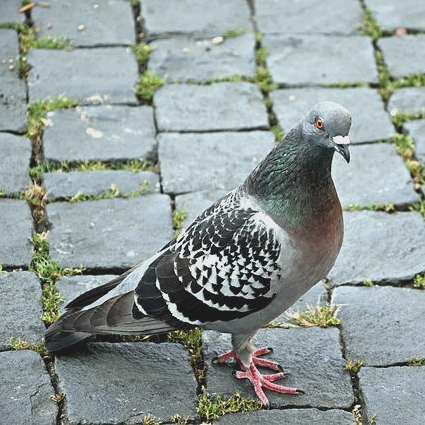}
\includegraphics[width=0.19\linewidth]{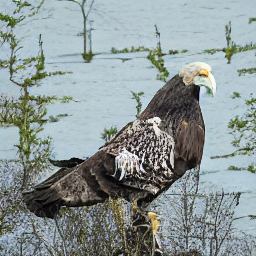}
\includegraphics[width=0.19\linewidth]{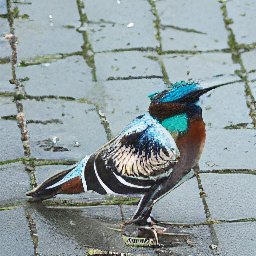}
\includegraphics[width=0.19\linewidth]{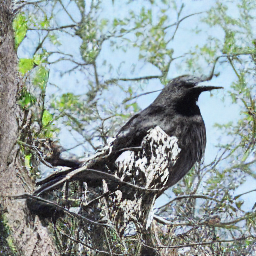}
\includegraphics[width=0.19\linewidth]{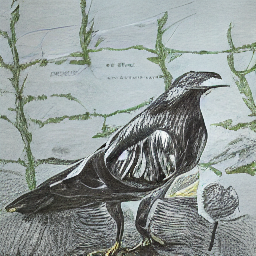}\\
\footnotesize{b) Editing global semantic attributes}\vspace{2pt}\\
\scriptsize{\enskip Input\quad\enskip Photograph\quad van Gogh\quad Picasso\quad Munch}\\
\includegraphics[width=0.19\linewidth]{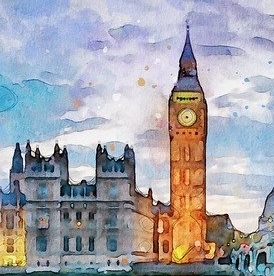}
\includegraphics[width=0.19\linewidth]{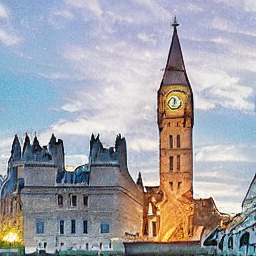}
\includegraphics[width=0.19\linewidth]{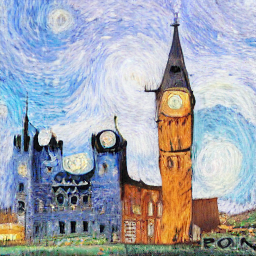}
\includegraphics[width=0.19\linewidth]{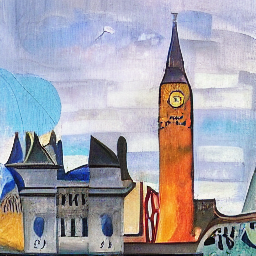}
\includegraphics[width=0.19\linewidth]{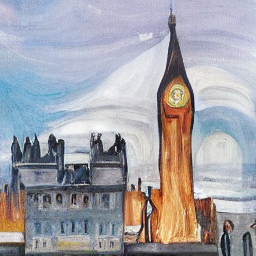}
\footnotesize{d) Artistic style transfer from text prompts }\vspace{2pt}\\
\end{minipage}
\vspace{1em}
\caption{LDEdit can  edit local and global semantic attributes and also perform  artistic style transfer on  real-world images using a single model. \label{fig:teaser}}
\end{figure}

In this work, we attempt to develop a fast and flexible approach to open domain image manipulation using arbitrary text prompts. Our goal is to accomplish a wide range of manipulations from text prompts  ranging from simple change in colour of an object, to  modification of multiple semantic attributes of image, and artistic styles,  all with a single model. 
Our work is inspired by the recent dramatic developments in  realistic image generation with  language guidance~\cite{ramesh2021zero,ding2021cogview,rombach2022high,ramesh2022hierarchical}. In particular, we leverage the recently proposed Latent Diffusion Model~(LDM) \cite{rombach2022high} which performs diffusion in a smaller dimensional latent space of  trained convolutional auto-encoders, to provide higher inference speed and computational efficiency. Further, we utilize the idea of non-Markovian diffusion proposed in Denoising Diffusion Implicit Models (DDIM)~\cite{song2020denoising}  which can enable faster inference and high fidelity sample reconstruction. Our key idea is the use of a shared latent representation as a link between the source image and the desired target. To this end, we employ a deterministic DDIM sampling in the forward diffusion in the latent space of LDM.  We use the same latent code along with the target text prompt to condition the reverse diffusion process, effectively achieving desired transformation in the input image, while automatically maintaining consistency with the original content representation. Using this technique, we can accomplish a variety of image manipulation tasks using the pretrained LDM, in a zero-shot fashion without further optimization or fine-tuning.  Further, by introducing controlled stochasticity, we can trade-off diversity for fidelity with original image. This is especially useful when the desired target is very different from the original input. 

Fig~\ref{fig:teaser} illustrates the diverse image editing tasks that can be accomplished by our LDEDit using only text prompts. We can modify objects in the image while largely preserving the original pose or structure, see Fig.~\ref{fig:teaser}~b). LDEdit can accomplish simultaneous global style manipulation as well as fine-grained (multiple) attribute changes such as change in expression, wrinkles, makeup while preserving identity in human faces, see Fig.~\ref{fig:teaser}~a). Further, without requiring an input mask, simple local edits such as adding a flower on a woman's hat, or eye glasses are achieved though text alone. Our approach can operate on diverse types of input images such as natural photographs, paintings, sketches, and strokes. By  providing an intuitive target text prompt " a photograph of a woman" or a "pixar animation of a woman", our method can translate from stroke to a semantically consistent image in the corresponding domain, see Fig.~\ref{fig:teaser}~c).  We can observe realistic details are hallucinated while transferring to the domain of natural photos,  for example, wrinkles in the picture of old man, in Fig.~\ref{fig:teaser}~c), or details in the clock  Fig.~\ref{fig:teaser}~d). Further, artistic style transfer is also achieved via simple text prompts, such as "a Picasso style painting". It can be seen that our approach can accomplish manipulations that are semantically and stylistically consistent with the given target text prompt, while remaining faithful to original content.  

By offering significant advantages in flexibility, faster run-times and  capability to generate diverse samples in parallel, LDEdit can facilitate efficient user-guided editing. Our experimental results demonstrate that LDEdit can accomplish diverse manipulation tasks, in addition to achieving performance close to recent state of the art baselines.
\begin{table}[t]
    \centering
    \resizebox{\linewidth}{!}{\begin{tabular}{ccccccc}
    \hline
        Method & Image input & Text input&Semantic & Artistic Style &  local edits & Comments  \\
         &  & &(Global) & &   & \\
       \hline
        DiffusionCLIP\cite{kim2021diffusionclip}&Class-Specific &Predefined & \ding{51} & \ding{51}&\ding{55}& Separately fine-tuned models for each task\\ 
        StyleCLIP\cite{patashnik2021styleclip}&Class-specific&Arbitrary& \ding{51} & \ding{55}&\ding{55}&Includes versions with and without optimization\\
        GLIDE\cite{nichol2021glide}&Open domain&Arbitrary& \ding{55} & \ding{55}&\ding{51}&Trained model for inpainting with mask input\\
        CLIPStyler\cite{kwon2021clipstyler}& Open domain&Arbitrary& \ding{55} & \ding{51}&\ding{55}&Test-time optimization w/o pretrained generator\\
        VQGAN+CLIP\cite{crowson2022vqgan}& Open domain&Arbitrary& \ding{51} & \ding{51}&Limited&Optimization with pretrained generator\\
         LDEdit(Ours)& Open-domain&Arbitrary& \ding{51} & \ding{51}&\ding{51}&A pretrained LDM is used\\
         \hline
    \end{tabular}}
    \vspace{0.5em}
    \caption{Comparison of recent state of the methods for text guided image manipulation. \label{tab:compare_methods}}
\end{table}
\section{Related Work}\label{sec:Related}
 \textbf{Image Generative Models~} 
Ever since the seminal works of VAEs~\cite{kingma2013auto} and  GANs~\cite{goodfellow2014generative}, image generative models have achieved significant improvements, and modern generative models can generate highly photo-realistic images \cite{brock2018large,vqvae2,karras2020analyzing,karras2021alias,esser2021taming,dhariwal2021diffusion,song2020denoising}. While GANs~\cite{goodfellow2014generative} achieve high quality generation, they are difficult to train and are prone to mode collapse. Likelihood-based models, \cite{kingma2013auto,vqvae2} on the other hand, have a stable training and capture more diversity. Score based~\cite{song2019generative,song2020score} or denoising diffusion~\cite{ho2020denoising,sohl2015deep}  models  are a new class of likelihood-based models built from a hierarchy of denoising auto-encoders~\cite{vincent2008extracting}. These models have recently demonstrated  generative capabilities surpassing GANs \cite{dhariwal2021diffusion,nichol2021improved}. Yet, high quality diffusion models are computationally expensive to train, and have slower inference times than GANs, due to expensive Markovian sampling and  iterative network evaluations required for diffusion. These problems can be alleviated by accelerated stochastic sampling techniques or by performing diffusion in a smaller latent space \cite{rombach2022high, vahdat2021score}. Employing deterministic diffusion process~\cite{song2020denoising} can also speed up inference,  in addition to enabling high fidelity sample reconstruction, which can be exploited for image recovery and manipulation.\\
\textbf{Image Manipulation~} As images can be manipulated in various ways, (\eg artistic style, image translation, semantic manipulation, local edits),  a variety of methods exist. Approaches for image translation include CNN based optimization using style and content images~\cite{gatys2016image}, conditional GANs trained on pair of domains~\cite{isola2017image,zhu2017unpaired,almahairi2018augmented,zhao2020unpaired}, GANs for multi-domain translation \cite{choi2018stargan,choi2020stargan} and more recently, conditional diffusion models~ \cite{sasaki2021unit,saharia2021palette}. An alternate approach \cite{zhu2016generative,brock2016neural} is to manipulate images in the latent space of pretrained GANs. StyleGANs~\cite{karras2020analyzing,karras2021alias} are a popular choice for such latent space editing due to their disentanglement properties in the latent space \cite{collins2020editing, shen2020interpreting,zhu2020domain, abdal2020image2stylegan++,gu2020image,wu2021stylespace}. This is achieved through optimization or by using encoders for GAN inversion~\cite{tov2021designing, richardson2021encoding, alaluf2021restyle}. However, GAN inversion may not yield faithful reconstruction \cite{bau2019seeing}. Improving StyleGAN inversion for editing is an active area of research~\cite{tov2021designing,alaluf2021restyle,dinh2021hyperinverter,wang2021HFGI,alaluf2022hyperstyle}. In contrast to GANs, diffusion models can readily be leveraged for inpainting~\cite{lugmayr2022repaint} and stroke guided image editing~\cite{meng2021sdedit} and even unpaired image translation~\cite{su2022dual}.\\
\textbf{Text Guided Generation and Manipulation:} Earlier works  employed RNNs   \cite{mansimov2015generating} and GANs \cite{reed2016generative, zhang2017stackgan,xu2018attngan, zhang2017stackgan,zhang2018stackgan++,dmgan, li2019controllable, zhang2021cross, zhu2022label} for text guided image synthesis, and manipulation \cite{dong2017semantic,li2020manigan,nam2018text}. Nevertheless, these works are often restricted to class specific image generation and are trained on smaller datasets. 
In the recent past, there is a rapid surge in vision-language models, with the developments in cross-modal contrastive learning~\cite{radford2021learning,jia2021scaling} and  powerful text-to-image generative models~\cite{ramesh2021zero, nichol2021glide, ramesh2022hierarchical,saharia2022photorealistic}. These models are trained on massive datasets to learn joint image-text distributions.
Some of these models \cite{ramesh2021zero,ding2021cogview,gafni2022make} use autoregressive(AR) transformers for generation, while  some others 
\cite{nichol2021glide,ramesh2022hierarchical,saharia2022photorealistic} employ diffusion based models for the generation task. However, training these models for high quality generation requires massive computational resources. To address this, some recent works  \cite{gu2021vector,tang2022improved,rombach2022high,bond2021unleashing,esser2021imagebart, Hu_2022_CVPR} instead perform the diffusion in a lower dimensional latent space resulting in faster training and inference.  In our work, we exploit Latent Diffusion Models (LDM) \cite{rombach2022high} as they offer good reconstruction quality, latency,  and perform diffusion in a continuous latent space.\\
\indent CLIP~\cite{radford2021learning} is a cross modal encoder which provides a similarity score between an image and a caption. Several recent approaches to text guided image synthesis~\cite{galatolo2021,advadnoun2021,crowson2022vqgan,clip_guided_diffusion,liu2021fusedream, liu2021more,couairon2022flexit,paiss2022no}  steer pretrained generative  models \cite{brock2018large, esser2021taming, dhariwal2021diffusion} towards a user provided text prompts using CLIP.  
This approach of  CLIP  controlled latent space navigation is directly applicable for image manipulation~\cite{crowson2022vqgan}, mask guided local editing~\cite{bau2021paint,avrahami2021blended},  semantic manipulation of class-specific images~\cite{patashnik2021styleclip,yu2022-CFCLIP,clip2stylegan} via StyleGAN inversion~\cite{alaluf2021restyle}. CLIP has also been applied to fine-tune  output domain and style~\cite{gal2021stylenada,  kim2021diffusionclip} of class-specific image generators.  While these approaches are promising,  optimization in latent space  for each text-prompt is expensive and time-consuming. On the other hand, the fine-tuned models are  fast, but restricted to the specific fine-tuned tasks. Further, class-specific generators are not suited for manipulation of open domain images. Instead of using pretrained generative models, some recent works employ  test-time optimization for each image and target text, using CLIP, for  tasks such as local object appearance~\cite{Text2LIVE2022}, global texture-style manipulation~\cite{kwon2021clipstyler}, rendering drawings~\cite{frans2021clipdraw,chendiffvg+}, however such optimization is task specific, and is expensive requiring many augmentations. Tab.~\ref{tab:compare_methods} provides an overview comparing the pros and cons of recent methods for text guided manipulation. As we can see, our approach and VQGAN+CLIP~\cite{crowson2022vqgan} can accomplish flexible manipulation tasks. Additionally, our approach allows fast manipulations. 
\section{Preliminaries}
\textbf{Diffusion Models:~}
Denoising diffusion probabilistic models~(DDPM)~\cite{ho2020denoising} are characterized by two diffusion processes: i)~a forward process to gradually corrupt  data samples into a tractable distribution e.g. Gaussian distribution, ii)~a learned iterative denoising process to convert Gaussian noise to samples from data distribution. 
 The forward diffusion involves progressively noising a clean image $x_0$ in $T$ time-steps with  transitions $\small{q(x_t \mid x_{t-1}) :=\mathcal{N}(\sqrt{1-\beta_t}x_{t-1}, \beta_t\mathbf{I})}$, where $\small{\{\beta_t\}^T_{t=0}}$  is the noise variance schedule.  The evolution of $\small{x_t}$ can be expressed as \vspace{-1pt}
\small
\begin{gather} 
\label{eq:forward_ddpm}
    x_t = \sqrt{\alpha_t}x_0  + \sqrt{(1 - \alpha_t)}\zeta,  \text{\quad\normalsize{where}\enskip}\zeta \sim \mathcal{N} (\mathbf{0,I}) \text{\enskip\normalsize{and}\enskip}\alpha_t := \prod_{s=1}^{t} {(1-\beta_s)}.
\end{gather}
\normalsize
The  generative process progressively denoises  $x_T$ to $x_0$ also via a Gaussian transition which is approximated by learned network $\epsilonb_\theta$. \\The reverse diffusion process is expressed as:
\small
\begin{gather}
    \label{eq:reverse_ddpm}
    x_{t-1} = \frac{1}{\sqrt{1-\beta_t}}\left(x_t - \frac{\beta_t}{\sqrt{1-\alpha_t}} \epsilonb_\theta(x_t, t)\right) + \sigma_t\xi, \text{\quad\normalsize where\enskip}\xi \sim \mathcal{N} (\mathbf{0,I}).
\end{gather}
\normalsize
\begin{figure}[t]
\centering
\scriptsize
\begin{minipage}[c]{0.6\textwidth}
\centering
\includegraphics[width=\linewidth]{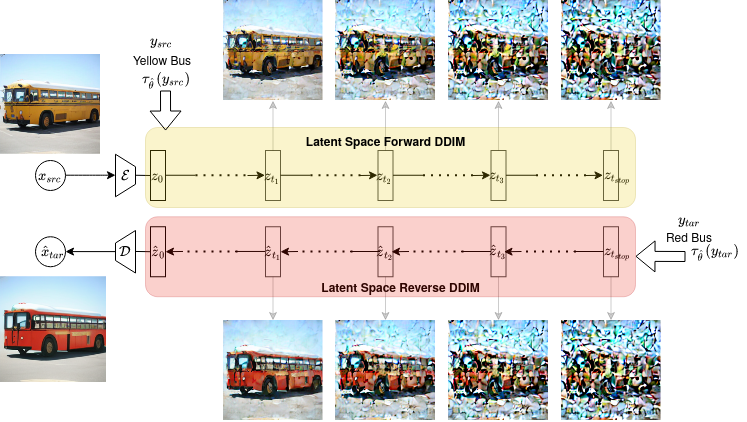}\\
a)~Overview of image manipulation using LDM.
\end{minipage}
\begin{minipage}[c]{0.33\textwidth}
\centering
\resizebox{\linewidth}{!}{\begin{tabular}{c}
$t_{stop} = 540$\qquad$t_{stop}=600$\qquad$t_{stop}=640$\\
   \includegraphics[width=0.33\linewidth]{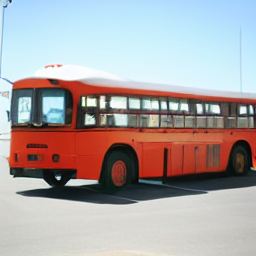}
\includegraphics[width=0.33\linewidth]{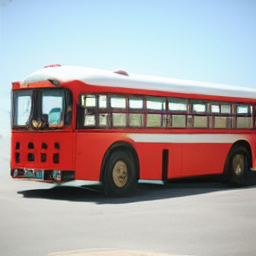}
\includegraphics[width=0.33\linewidth]{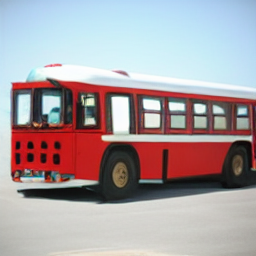}\\
b) Effect of varying $t_{stop}$  ($\eta=0$)\vspace{2pt}\\
  \hspace{-5pt}\begin{sideways}
  $\eta=0.3$
  \end{sideways} \includegraphics[width=0.33\linewidth]{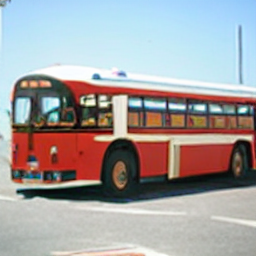}
\includegraphics[width=0.33\linewidth]{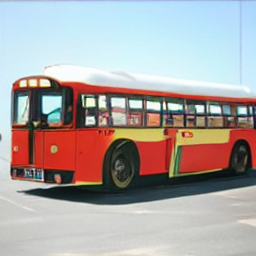}
\includegraphics[width=0.33\linewidth]{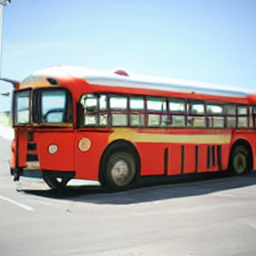}\\
  \hspace{-5pt}\begin{sideways}
  $\eta=0.6$
  \end{sideways}
   \includegraphics[width=0.33\linewidth]{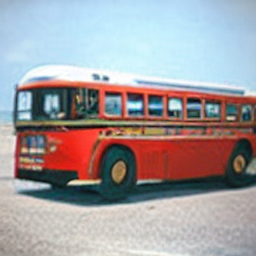}
\includegraphics[width=0.33\linewidth]{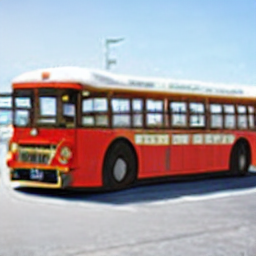}
\includegraphics[width=0.33\linewidth]{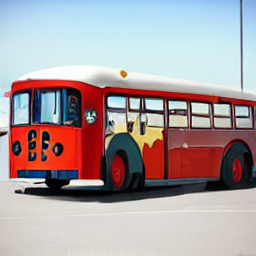}\\
c) Effect of varying $\eta$, with $t_{stop}=540$

\end{tabular}}
\end{minipage}\vspace{1em}
\caption{a) Overview of LDEdit, illustrating forward and reverse diffusion in latent space of autoencoder. b) and c) illustrate the effects of varying time steps $t_{stop}$ and stochasticity hyperparameter $\small{\eta}$ respectively \label{fig:method}}

\end{figure}
\noindent Denoising Diffusion Implicit Models(DDIM) ~\cite{song2020denoising} employ a different non-Markovian forward process with the same forward marginals as DDPM:
\small
\begin{gather}
x_{t-1} = \sqrt{\alpha_{t-1}}\left(\frac{x_{t} - \sqrt{1-\alpha_t}\epsilonb_{\theta}(x_t,t)}{\sqrt{\alpha_{t}}}\right) +  \sqrt{1 - \alpha_{t-1} - \sigma^2_t}{\epsilonb}_{\theta}(x_{t}, t) + \sigma^2_t\xi, 
\label{eq:ddim_original}
\end{gather}
\normalsize 
where  $\small{\xi \sim \mathcal{N} (\mathbf{0,I})}$ and $\small{\alpha_0 := 1}$, by definition. 
Varying $\small{\sigma}$ leads to different generative processes with the same model $\epsilonb_\theta$. When $\small{\sigma_t}$ is set to 0, the DDIM sampling becomes fully deterministic, enabling fast inversion of the noised latent variable to the original images~(or to $x_0$ in our case)  \cite{song2020denoising,song2020score}. In this case, the deterministic 
forward DDIM process expressed as: 
\small
\begin{gather}
\label{eq:forward_ddim}
x_{t+1} = \sqrt{\alpha_{t+1}}\left(\frac{x_{t} - \sqrt{1-\alpha_t}\epsilonb_{\theta}(x_t,t)}{\sqrt{\alpha_{t}}}\right) +  \sqrt{1 - \alpha_{t+1}}\bm{\epsilon}_{{\theta}}(x_{t}, t) 
\end{gather}
\normalsize
and the deterministic reverse DDIM process is expressed as:
\small
\begin{gather}
\label{eq:reverse_ddim}
x_{t-1} = \sqrt{\alpha_{t-1}}\left(\frac{x_{t} - \sqrt{1-\alpha_t}\epsilonb_{\theta}(x_t,t)}{\sqrt{\alpha_{t}}}\right) +  \sqrt{1 - \alpha_{t-1}}\bm{\epsilon}_{\theta}(x_{t}, t) 
\end{gather}
 \normalsize
For different subsequences $\tau$ in $[1, \ldots, T]$ \cite{song2020denoising} consider $\small{\sigma}$ of the form:
\small
\begin{gather}
    \sigma_{\tau_{i}}(\eta) = \eta \sqrt{(1 - \alpha_{\tau_{i-1}})/(1 - \alpha_{\tau_{i}})}\sqrt{1 - {\alpha_{\tau_i}}/{\alpha_{\tau_{i-1}}}}, 
    \label{eq:eta}
\end{gather}
\normalsize
where the hyperparameter $\small{\eta \in \mathbb{R}_{\geq 0}}$ controls the degree of stochasticity, with $\small{\eta = 1}$ leading to original DDPM generative process and $\small{\eta = 0}$ leading to DDIM.\vspace{2pt}\\
\textbf{Latent Diffusion Models:~}
The main idea of LDMs is to perform diffusion in the latent space of an autoencoder to improve speed and computational efficiency.
Given an image $\small{x_{\mbox{src}} }\in \mathbb{R}^{H \times W \times C}$, 
the encoder $\mathcal{E}$ maps $\small{x_{\mbox{src}} }$ into a down-sampled latent code $z_0=\mathcal{E}(\small{x_{\mbox{src}} })$, and the decoder $\mathcal{D}$ is trained to 
recover the image from this latent. 
This encoding results in a lossy compression, i.e. $\|\mathcal{D}(\mathcal{E}(\small{x_{\mbox{src}} }))-\small{x_{\mbox{src}} }\|$ is finite, which is a trade-off for computational efficiency. Following encoding into latent space, diffusion process can happen via DDPM or  DDIM \eqref{eq:forward_ddpm}$-$\eqref{eq:reverse_ddim}, but in $z_t$ for $t\in[1,T]$ instead of $x_t$. The diffusion process can  additionally be  conditioned on user inputs such as text prompts $\small{\epsilonb_{\theta}(z_t,t,\tau_{\tilde{\theta}}(y))}$. Here, the text-prompts $y$ are tokenized using  transformers  $\small{\tau_{\tilde{\theta}}}$ \cite{vaswani2017attention} for conditioning the diffusion process.
\begin{figure}[t]
\scriptsize
\centering
Input: Yellow bus $->$ Target: Tram\\
\includegraphics[width=0.095\linewidth]{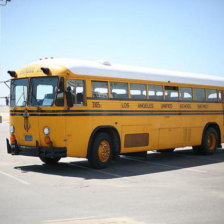}
\includegraphics[width=0.095\linewidth]{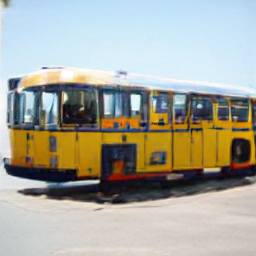}
\hspace{-1.6pt}\includegraphics[width=0.095\linewidth]{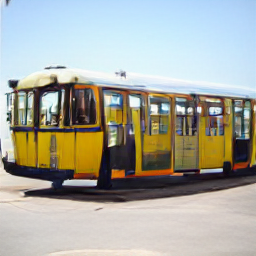}
\includegraphics[width=0.095\linewidth]{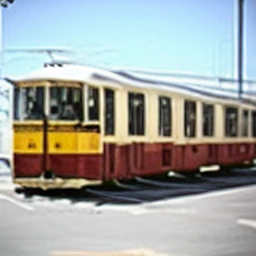}
\includegraphics[width=0.095\linewidth]{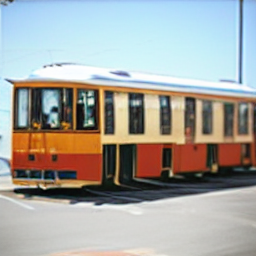}
\hspace{-1.6pt}\includegraphics[width=0.095\linewidth]{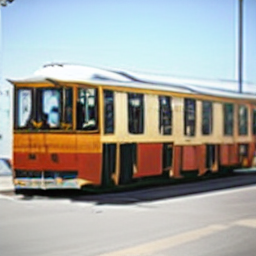}
\hspace{-1.6pt}\includegraphics[width=0.095\linewidth]{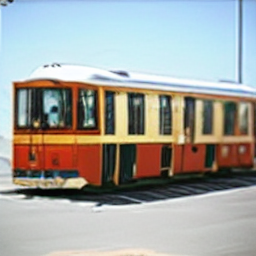}
\includegraphics[width=0.095\linewidth]{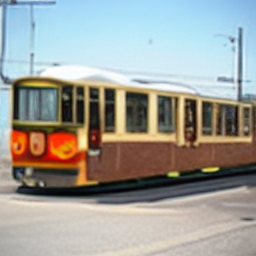}
\hspace{-1.6pt}\includegraphics[width=0.095\linewidth]{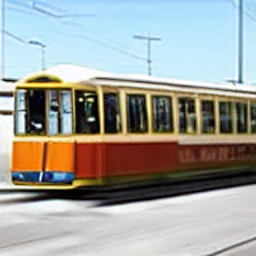}
\hspace{-1.6pt}\includegraphics[width=0.095\linewidth]{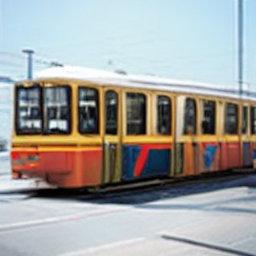}\\
Input: Yellow bus $->$ Target: Truck\\
\includegraphics[width=0.095\linewidth]{images/bus_compare/yellow_bus-red_bus.png}
\includegraphics[width=0.095\linewidth]{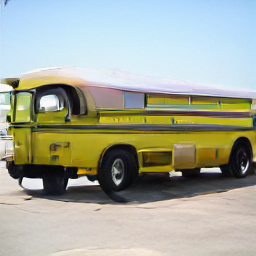}
\hspace{-1.6pt}\includegraphics[width=0.095\linewidth]{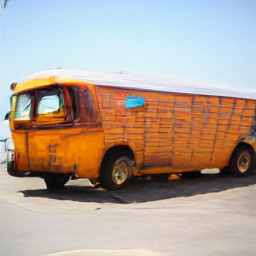}
\includegraphics[width=0.095\linewidth]{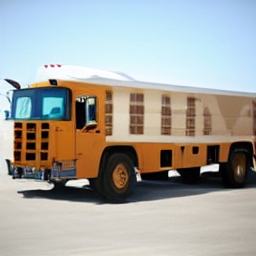}
\includegraphics[width=0.095\linewidth]{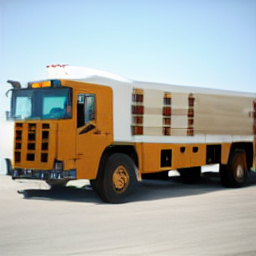}
\hspace{-1.6pt}\includegraphics[width=0.095\linewidth]{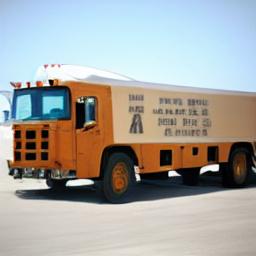}
\hspace{-1.6pt}\includegraphics[width=0.095\linewidth]{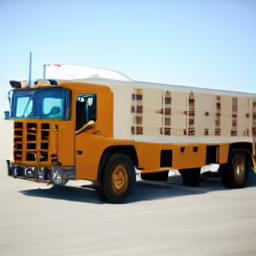}
\includegraphics[width=0.095\linewidth]{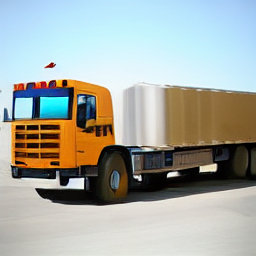}
\hspace{-1.6pt}\includegraphics[width=0.095\linewidth]{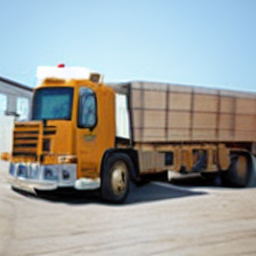}
\hspace{-1.6pt}\includegraphics[width=0.095\linewidth]{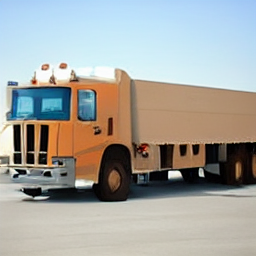}\\
Input: Yellow bus $->$ Target: Red steam engine\\
\includegraphics[width=0.095\linewidth]{images/bus_compare/yellow_bus-red_bus.png}
\includegraphics[width=0.095\linewidth]{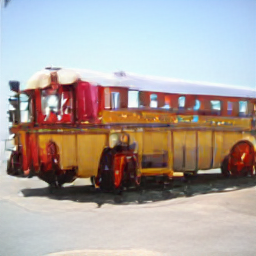}
\hspace{-1.6pt}\includegraphics[width=0.095\linewidth]{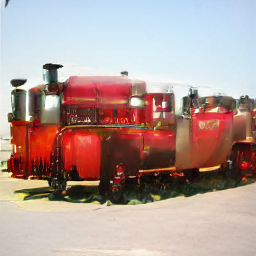}
\includegraphics[width=0.095\linewidth]{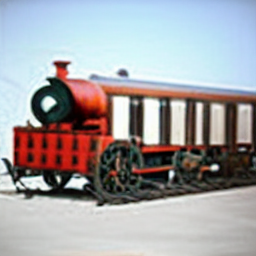}
\includegraphics[width=0.095\linewidth]{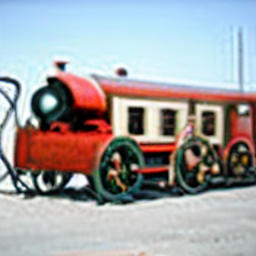}
\hspace{-1.6pt}\includegraphics[width=0.095\linewidth]{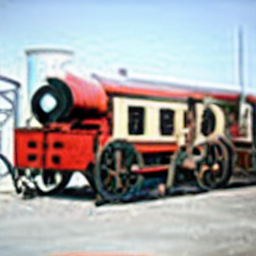}
\hspace{-1.6pt}\includegraphics[width=0.095\linewidth]{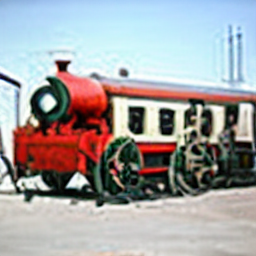}
\includegraphics[width=0.095\linewidth]{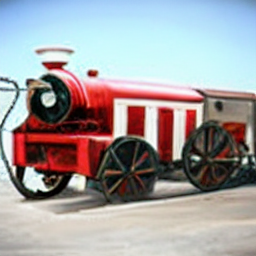}
\hspace{-1.6pt}\includegraphics[width=0.095\linewidth]{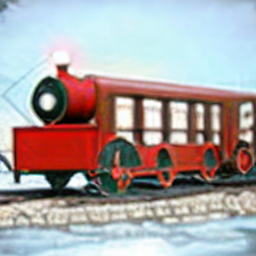}
\hspace{-1.6pt}\includegraphics[width=0.095\linewidth]{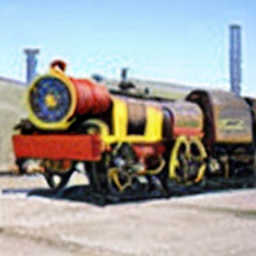}
\par
\quad\enskip Input\qquad\enskip VQGAN+CLIP \cite{crowson2022vqgan} \quad Ours $\eta=0$\quad$\xleftarrow{\hspace*{0.07\linewidth}} \text{ Ours }\eta = 0.3 \xrightarrow{\hspace*{0.07\linewidth}}$\enskip$\xleftarrow{\hspace*{0.07\linewidth}} \text{ Ours }\eta = 0.6 \xrightarrow{\hspace*{0.07\linewidth}}$\\
\vspace{1em}
\caption{Comparison with VQGAN+CLIP~\cite{crowson2022vqgan}: Manipulation results  of  yellow bus according to target texts `a tram', `a truck' and `a red steam engine'.\label{fig:bus}}
\end{figure}
\begin{figure}[t]
\scriptsize
\resizebox{\linewidth}{!}{\begin{tabular}{c}
\resizebox{\linewidth}{!}{
\begin{tabular}{l ccc ccc ccc}
Face&\quad Tanned&Zuckerberg&Pixar&\quad Tanned&Zuckerberg&Pixar&\quad Tanned&Zuckerberg&Pixar
\end{tabular}}\\
\includegraphics[width=0.1\linewidth]{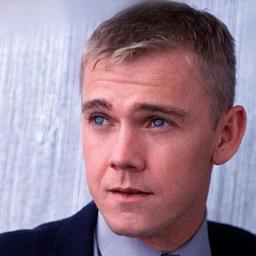}
\includegraphics[width=0.1\linewidth]{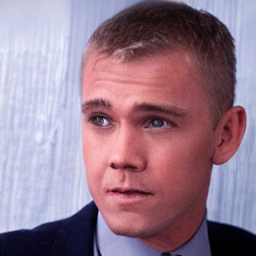}
\hspace{-1.6pt}\includegraphics[width=0.1\linewidth]{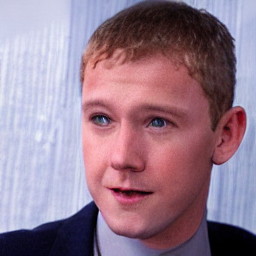}
\hspace{-1.6pt}\includegraphics[width=0.1\linewidth]{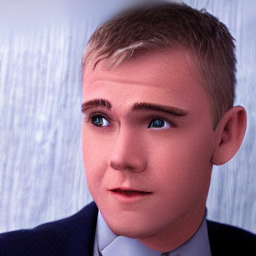}
\includegraphics[width=0.1\linewidth]{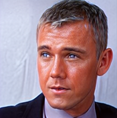}
\hspace{-1.6pt}\includegraphics[width=0.1\linewidth]{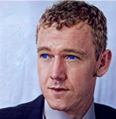}
\hspace{-1.6pt}\includegraphics[width=0.1\linewidth]{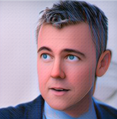}
\includegraphics[width=0.1\linewidth]{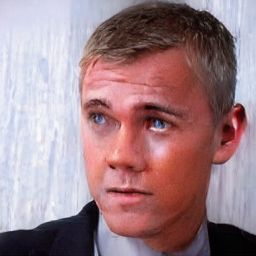}
\hspace{-1.6pt}\includegraphics[width=0.1\linewidth]{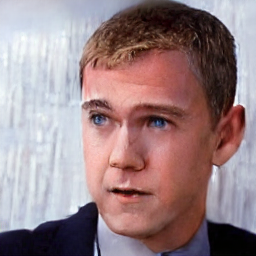}
\hspace{-1.6pt}\includegraphics[width=0.1\linewidth]{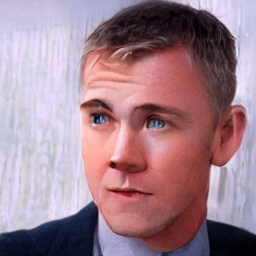}
\\
\resizebox{\linewidth}{!}{
\begin{tabular}{c ccc ccc ccc}
Dog&\quad Bear&\quad Fox&\enskip NicolasCage&Bear&Fox&NicolasCage&Bear&Fox&NicolasCage
\end{tabular}}\\
\includegraphics[width=0.1\linewidth]{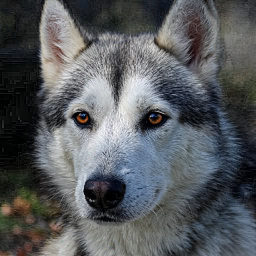}
\includegraphics[width=0.1\linewidth]{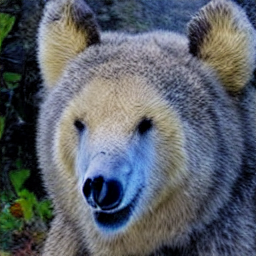}
\hspace{-1.6pt}\includegraphics[width=0.1\linewidth]{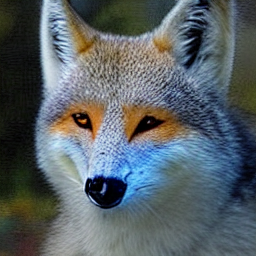}
\hspace{-1.6pt}\includegraphics[width=0.1\linewidth]{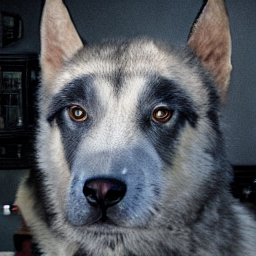}
\includegraphics[width=0.1\linewidth]{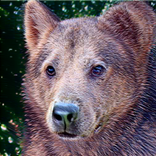}
\hspace{-1.6pt}\includegraphics[width=0.1\linewidth]{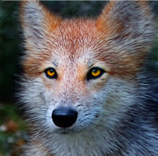}
\hspace{-1.6pt}\includegraphics[width=0.1\linewidth]{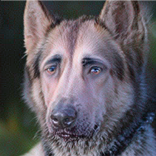}
\includegraphics[width=0.1\linewidth]{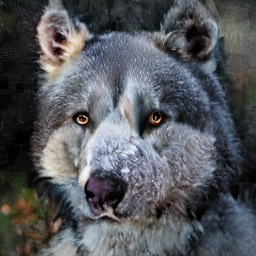}
\hspace{-1.6pt}\includegraphics[width=0.1\linewidth]{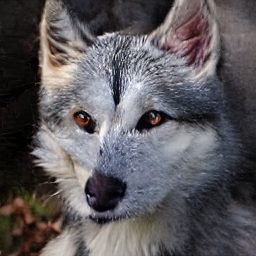}
\hspace{-1.6pt}\includegraphics[width=0.1\linewidth]{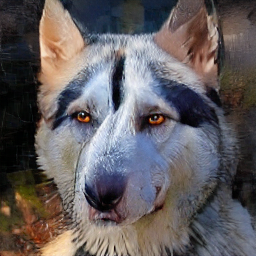}
\\
\resizebox{\linewidth}{!}{
\begin{tabular}{l ccc ccc ccc}
\hspace{-5pt}Tennisball&Baseball& Orange&Tomato&Baseball& Orange&Tomato&Baseball& Orange&Tomato
\end{tabular}}\\
\includegraphics[width=0.1\linewidth]{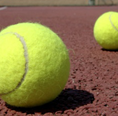}
\includegraphics[width=0.1\linewidth]{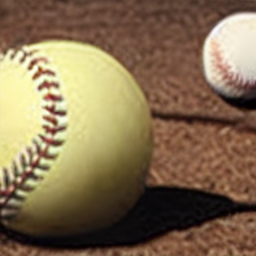}
\hspace{-1.6pt}\includegraphics[width=0.1\linewidth]{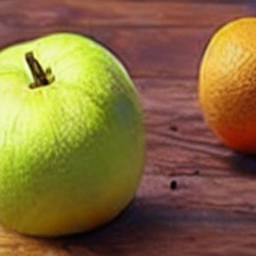}
\hspace{-1.6pt}\includegraphics[width=0.1\linewidth]{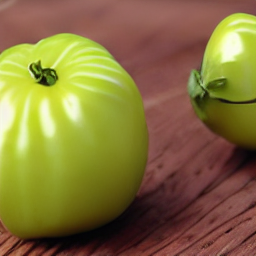}
\includegraphics[width=0.1\linewidth]{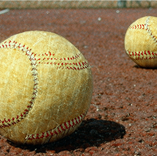}
\hspace{-1.6pt}\includegraphics[width=0.1\linewidth]{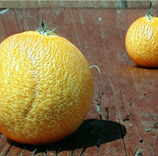}
\hspace{-1.6pt}\includegraphics[width=0.1\linewidth]{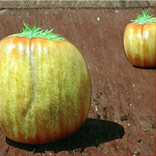}
\includegraphics[width=0.1\linewidth]{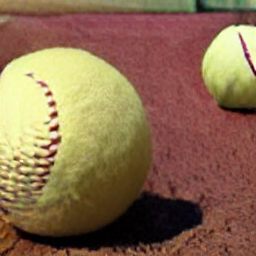}
\hspace{-1.6pt}\includegraphics[width=0.1\linewidth]{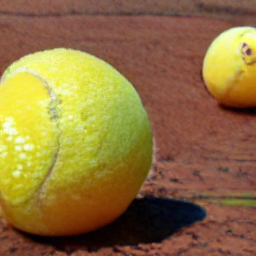}
\hspace{-1.6pt}\includegraphics[width=0.1\linewidth]{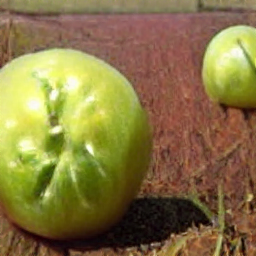}\\
\resizebox{\linewidth}{!}{
\begin{tabular}{l ccc ccc ccc}
Stroke&\enskip van Gogh&\enskip Pixar\enskip&Neanderthal&van Gogh& Pixar&Neanderthal&van Gogh& Pixar&Neanderthal
\end{tabular}}\\
\includegraphics[width=0.1\linewidth]{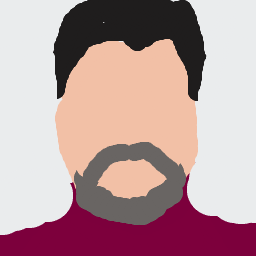}
\includegraphics[width=0.1\linewidth]{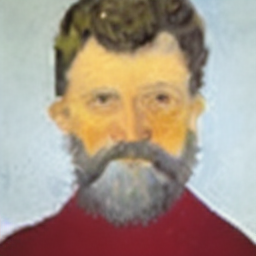}
\hspace{-1.6pt}\includegraphics[width=0.1\linewidth]{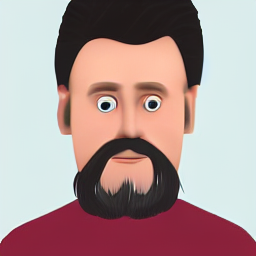}
\hspace{-1.6pt}\includegraphics[width=0.1\linewidth]{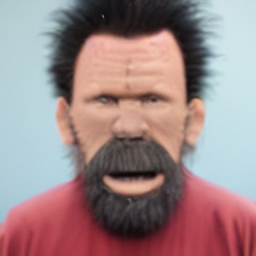}
\includegraphics[width=0.1\linewidth]{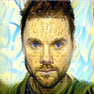}
\hspace{-1.6pt}\includegraphics[width=0.1\linewidth]{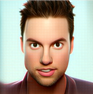}
\hspace{-1.6pt}\includegraphics[width=0.1\linewidth]{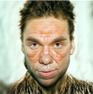}
\includegraphics[width=0.1\linewidth]{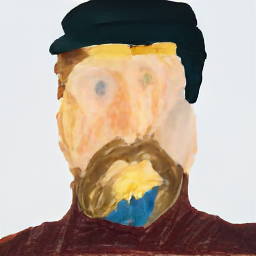}
\hspace{-1.6pt}\includegraphics[width=0.1\linewidth]{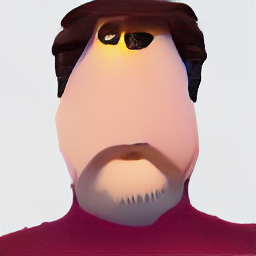}
\hspace{-1.6pt}\includegraphics[width=0.1\linewidth]{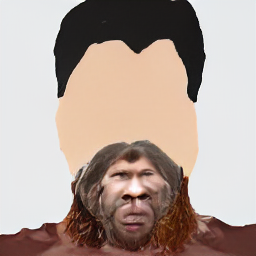}\\
\qquad Input\quad\enskip$\xleftarrow{\hspace*{0.07\linewidth}} \text{ LDEdit(Ours) } \xrightarrow{\hspace*{0.07\linewidth}}$\quad $\xleftarrow{\hspace*{0.05\linewidth}}\text{DiffusionCLIP \cite{kim2021diffusionclip}} \xrightarrow{\hspace*{0.05\linewidth}}$\enskip$\xleftarrow{\hspace*{0.05\linewidth}}\text{VQGAN+CLIP \cite{crowson2022vqgan}} \xrightarrow{\hspace*{0.05\linewidth}}$\\

\end{tabular}}
\vspace{1em}
\caption{Visual comparison of image manipulation task with DiffusionCLIP~\cite{kim2021diffusionclip} and VQGAN+CLIP~\cite{crowson2022vqgan}. Our LDEdit can successfully transform input image into target classes while retaining the original pose.\label{fig:dclip}}
\end{figure}
\begin{figure}[t]
\centering
\scriptsize
\resizebox{0.9\textwidth}{!}{\begin{tabular}{c}
\hspace{2em}Original\qquad Recon\qquad Watercolor\qquad Original\quad Neanderthal\quad Zombie\qquad Original\qquad Makeup\quad +Curly hair\vspace{-4pt}\\
\begin{sideways}\textbf{LDEdit}(Ours)\end{sideways}\includegraphics[width=0.1\linewidth]{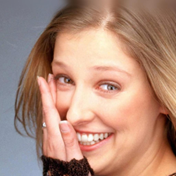}
\hspace{-2pt}\includegraphics[width=0.1\linewidth]{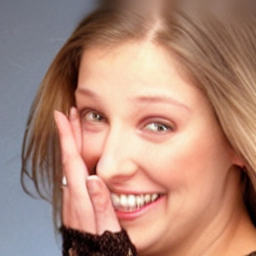}
\hspace{-2pt}\includegraphics[width=0.1\linewidth]{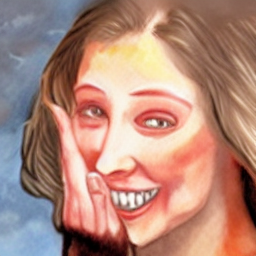}
~\includegraphics[width=0.1\linewidth]{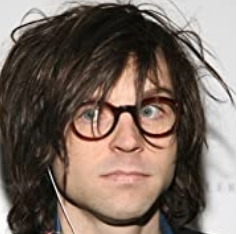}
\hspace{-2pt}\includegraphics[width=0.1\linewidth]{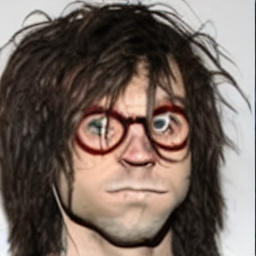}
\hspace{-2pt}\includegraphics[width=0.1\linewidth]{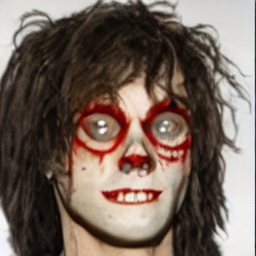}
~\includegraphics[width=0.1\linewidth]{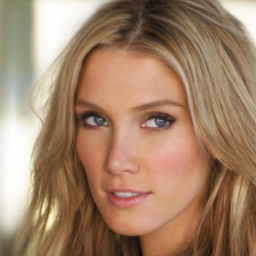}
\hspace{-2pt}\includegraphics[width=0.1\linewidth]{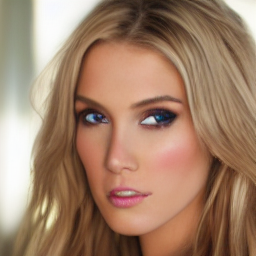}
\hspace{-2pt}\includegraphics[width=0.1\linewidth]{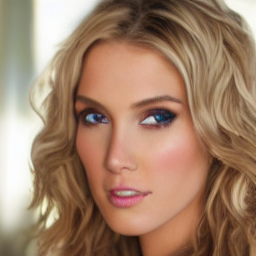}
\\
\begin{sideways}~DiffClip\end{sideways}\includegraphics[width=0.1\linewidth]{images/inputs/diffClip1.png}
\hspace{-2pt}\includegraphics[width=0.1\linewidth]{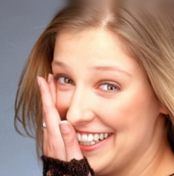}
\hspace{-2pt}\includegraphics[width=0.1\linewidth]{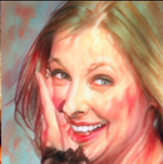}
~\includegraphics[width=0.1\linewidth]{images/imdb/imdb.png}
\hspace{-2pt}\includegraphics[width=0.1\linewidth]{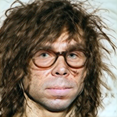}
\hspace{-2pt}\includegraphics[width=0.1\linewidth]{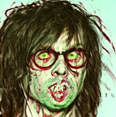}
~\includegraphics[width=0.1\linewidth]{images/inputs/celeb3.png}
\hspace{-2pt}\includegraphics[width=0.1\linewidth]{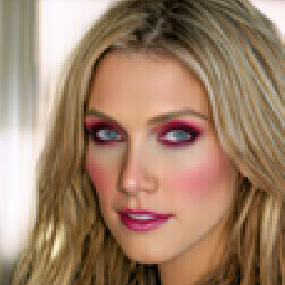}
\hspace{-2pt}\includegraphics[width=0.1\linewidth]{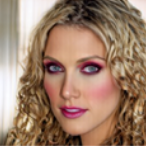}\\
\begin{sideways} ~~StyleClip\end{sideways}\includegraphics[width=0.1\linewidth]{images/inputs/diffClip1.png}
\hspace{-2pt}\includegraphics[width=0.1\linewidth]{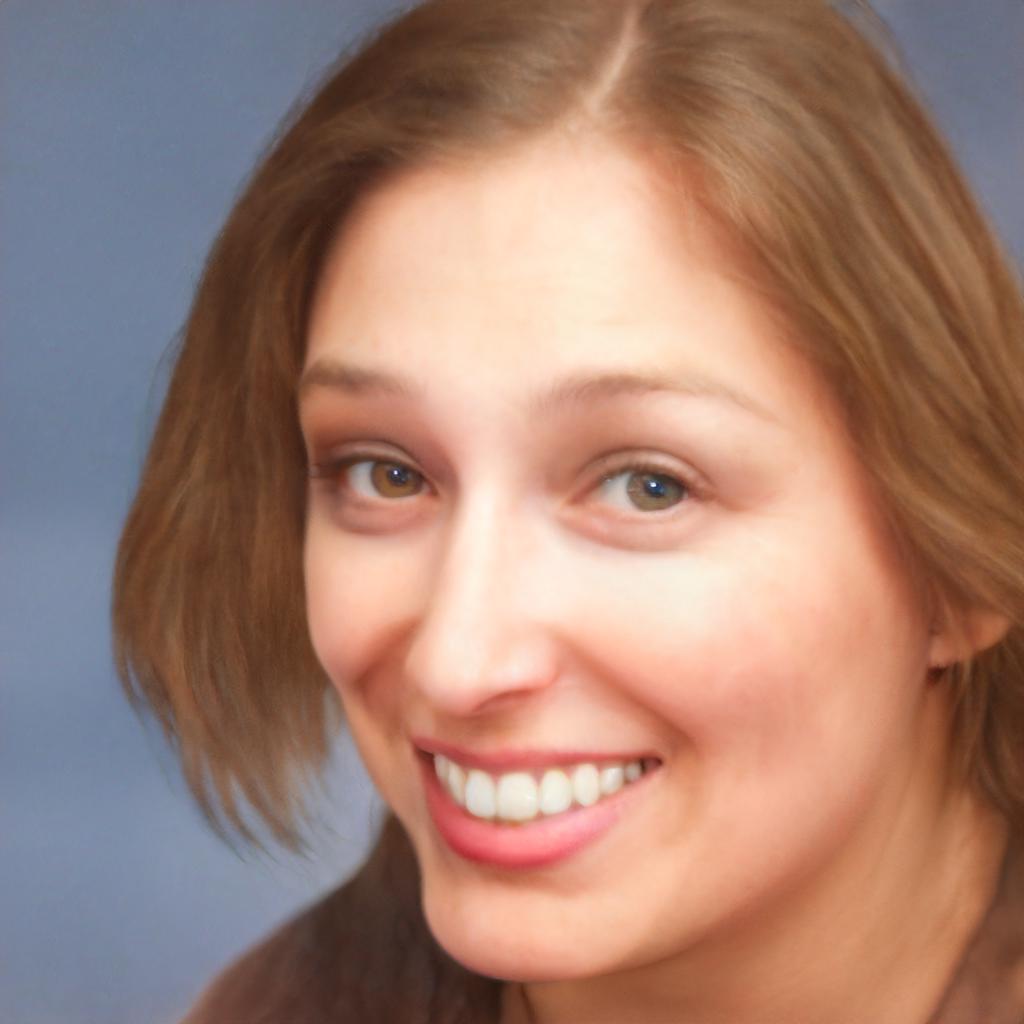}
\hspace{-2pt}\includegraphics[width=0.1\linewidth]{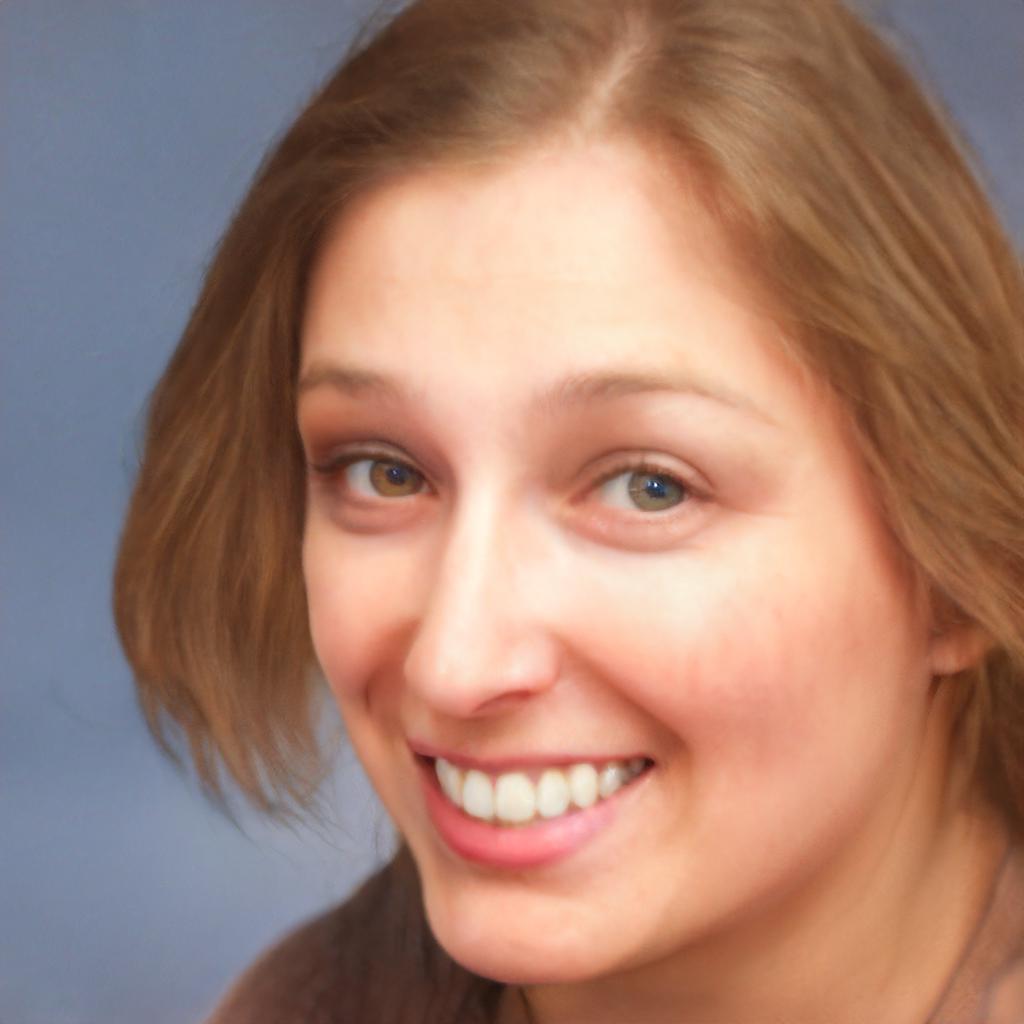}
~\includegraphics[width=0.1\linewidth]{images/imdb/imdb.png}
\hspace{-2pt}\includegraphics[width=0.1\linewidth]{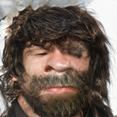}
\hspace{-2pt}\includegraphics[width=0.1\linewidth]{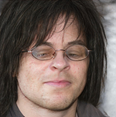}
~\includegraphics[width=0.1\linewidth]{images/inputs/celeb3.png}
\hspace{-2pt}\includegraphics[width=0.1\linewidth]{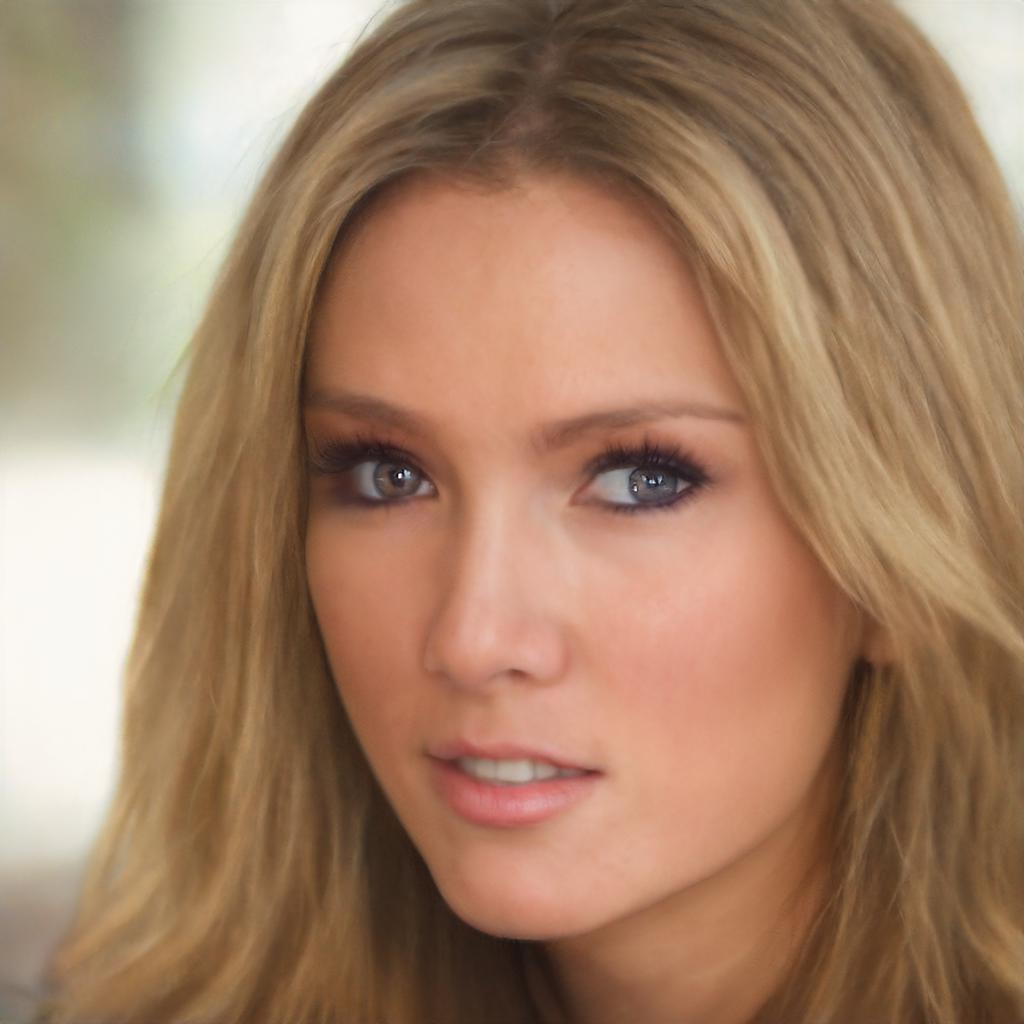}
\hspace{-2pt}\includegraphics[width=0.1\linewidth]{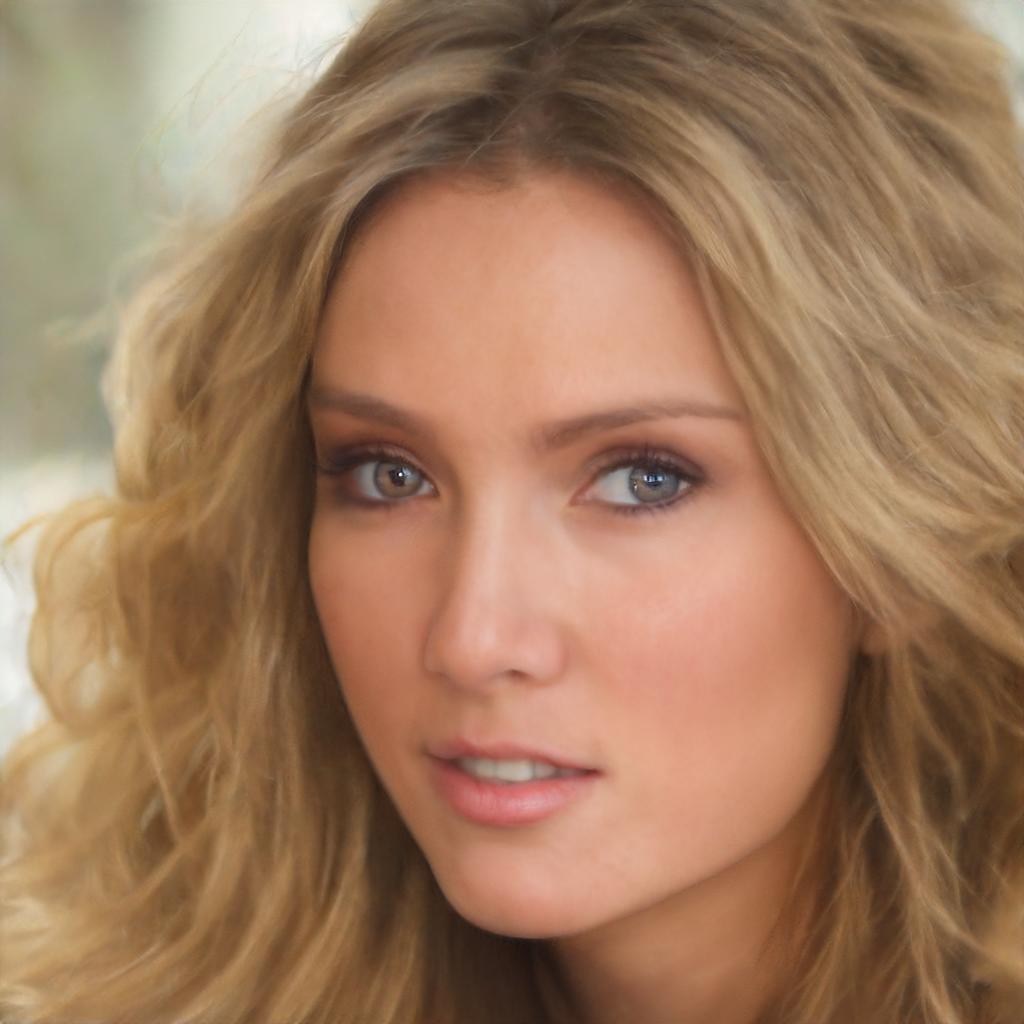}\\
\begin{sideways} Style-NADA\end{sideways}\includegraphics[width=0.1\linewidth]{images/inputs/diffClip1.png}
\hspace{-2pt}\includegraphics[width=0.1\linewidth]{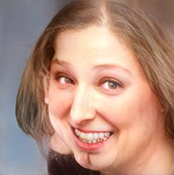}
\hspace{-2pt}\includegraphics[width=0.1\linewidth]{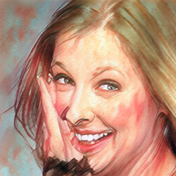}
~\includegraphics[width=0.1\linewidth]{images/imdb/imdb.png}
\hspace{-2pt}\includegraphics[width=0.1\linewidth]{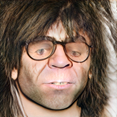}
\hspace{-2pt}\includegraphics[width=0.1\linewidth]{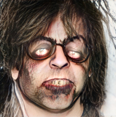}
~\includegraphics[width=0.1\linewidth]{images/inputs/celeb3.png}
\hspace{-2pt}\includegraphics[width=0.1\linewidth]{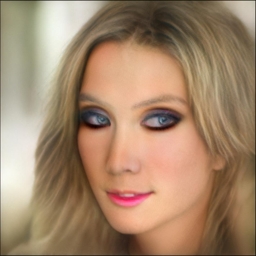}
\hspace{-2pt}\includegraphics[width=0.1\linewidth]{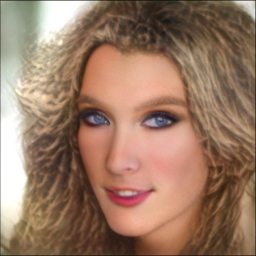}\\
\begin{sideways}~ TediGAN\end{sideways}~\includegraphics[width=0.1\linewidth]{images/inputs/diffClip1.png}
\hspace{-2pt}\includegraphics[width=0.1\linewidth]{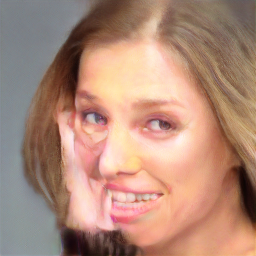}
\hspace{-2pt}\includegraphics[width=0.1\linewidth]{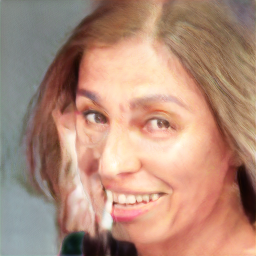}
~\includegraphics[width=0.1\linewidth]{images/imdb/imdb.png}
\hspace{-2pt}\includegraphics[width=0.1\linewidth]{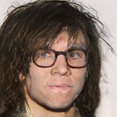}
\hspace{-2pt}\includegraphics[width=0.1\linewidth]{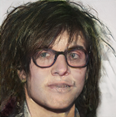}
~\includegraphics[width=0.1\linewidth]{images/inputs/celeb3.png}
\hspace{-2pt}\includegraphics[width=0.1\linewidth]{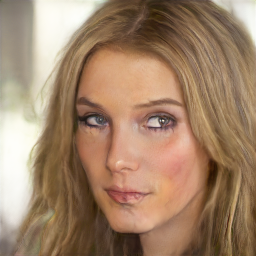}
\hspace{-2pt}\includegraphics[width=0.1\linewidth]{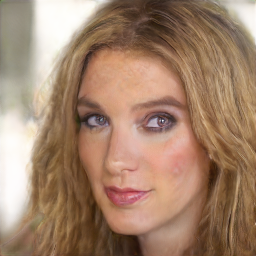}\\
\begin{sideways} VQGANClip\end{sideways}~\includegraphics[width=0.1\linewidth]{images/inputs/diffClip1.png}
\hspace{-2pt}\includegraphics[width=0.1\linewidth]{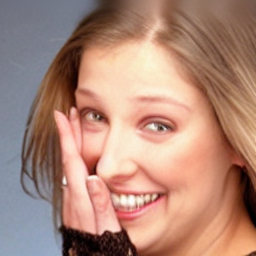}
\hspace{-2pt}\includegraphics[width=0.1\linewidth]{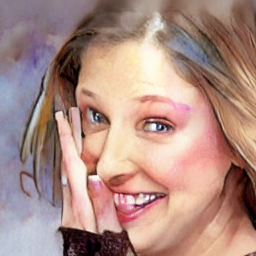}
~\includegraphics[width=0.1\linewidth]{images/imdb/imdb.png}
\hspace{-2pt}\includegraphics[width=0.1\linewidth]{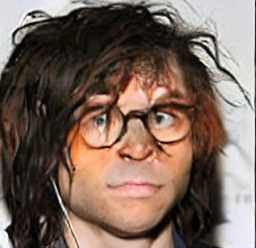}
\hspace{-2pt}\includegraphics[width=0.1\linewidth]{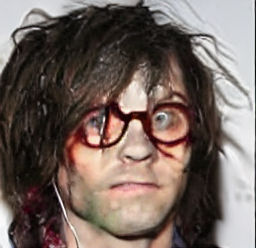}
~\includegraphics[width=0.1\linewidth]{images/inputs/celeb3.png}
\hspace{-2pt}\includegraphics[width=0.1\linewidth]{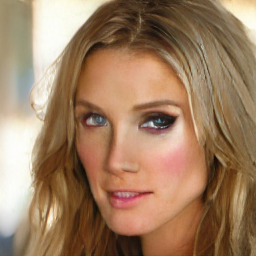}
\hspace{-2pt}\includegraphics[width=0.1\linewidth]{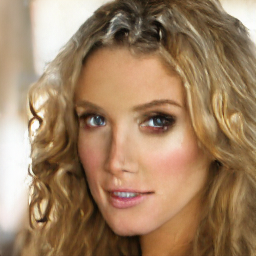}\\
\begin{sideways} ClipStyler\end{sideways}~\includegraphics[width=0.1\linewidth]{images/inputs/diffClip1.png}
\hspace{-2pt}\includegraphics[width=0.1\linewidth]{images/vqgan_clip_face_res/diffClip1_recon_no_text_nodiffn.png}
\hspace{-2pt}\includegraphics[width=0.1\linewidth]{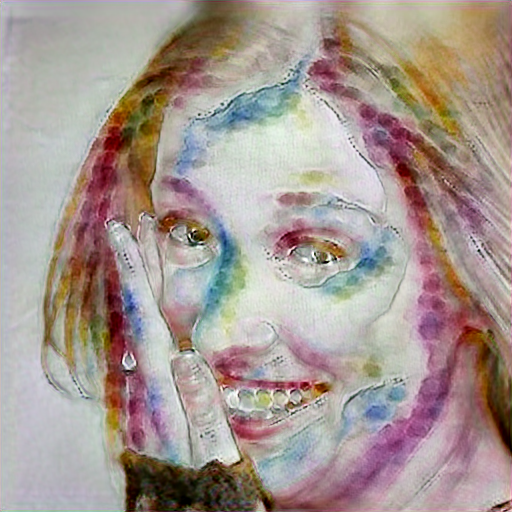}
~\includegraphics[width=0.1\linewidth]{images/imdb/imdb.png}
\hspace{-2pt}\includegraphics[width=0.1\linewidth]{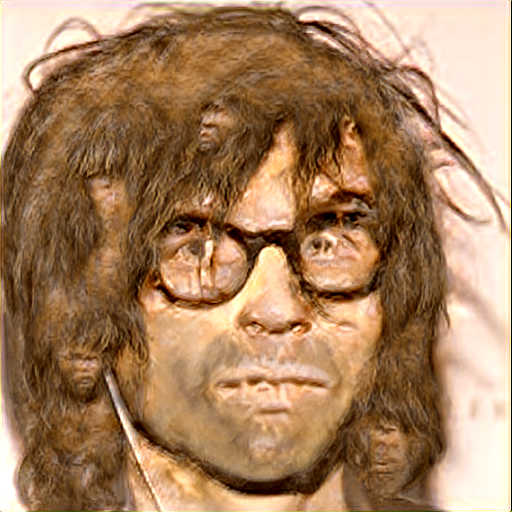}
\hspace{-2pt}\includegraphics[width=0.1\linewidth]{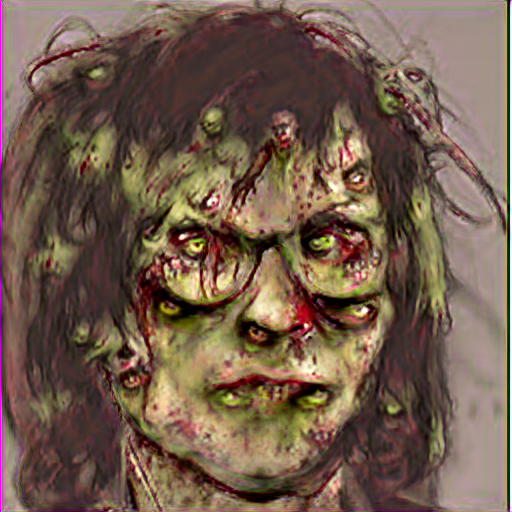}
~\includegraphics[width=0.1\linewidth]{images/inputs/celeb3.png}
\hspace{-2pt}\includegraphics[width=0.1\linewidth]{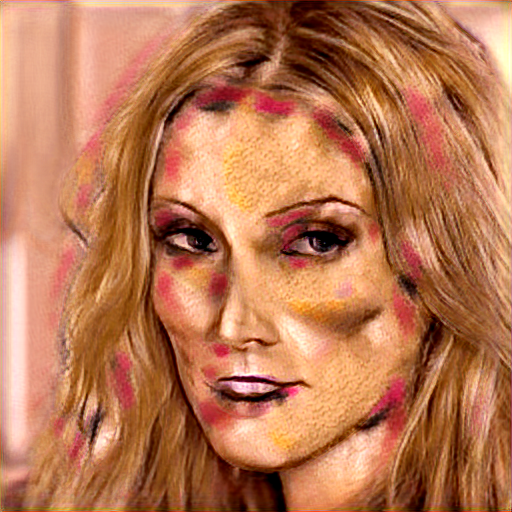}
\hspace{-2pt}\includegraphics[width=0.1\linewidth]{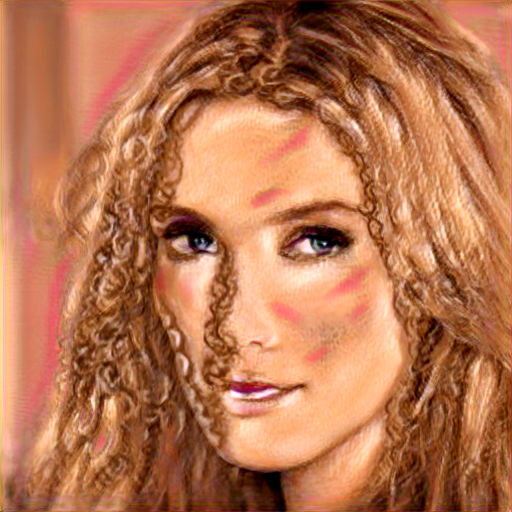}\\
  a) Reconstruction and style transfer\qquad\qquad\quad b) Domain transfer\qquad\qquad\quad c) Multi-attribute semantic changes\vspace{1em}\\
\end{tabular}}

\caption{Comparison with recent baselines: DiffusionCLIP\cite{kim2021diffusionclip}  StyleCLIP\cite{patashnik2021styleclip}, StyleGAN-NADA ~\cite{gal2021stylenada}, TEDIGAN~\cite{xia2021tedigan}, CLIPStyler~\cite{kwon2021clipstyler}, VQGAN+CLIP~\cite{crowson2022vqgan}. \label{fig:faces}}

\end{figure}
\begin{figure}[h]
\scriptsize
\centering
\resizebox{0.95\linewidth}{!}{\begin{tabular}{c}
\includegraphics[width=0.1\linewidth]{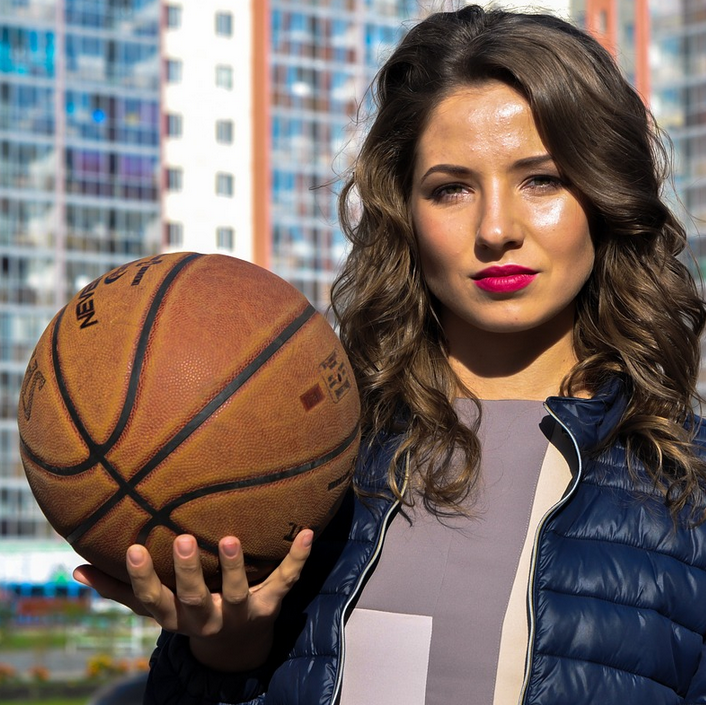}
\includegraphics[width=0.1\linewidth]{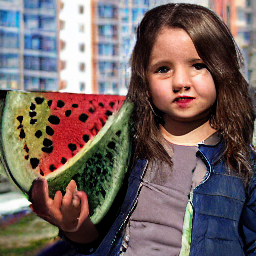}
\hspace{-1.2pt}\includegraphics[width=0.1\linewidth]{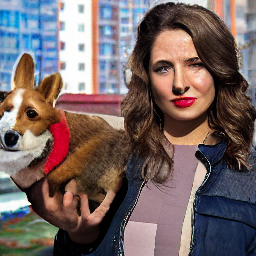}
\hspace{-1.2pt}\includegraphics[width=0.1\linewidth]{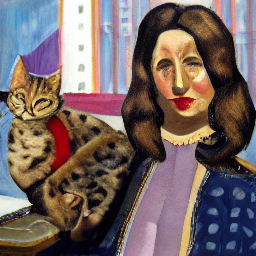}
\hspace{-1.2pt}\includegraphics[width=0.1\linewidth]{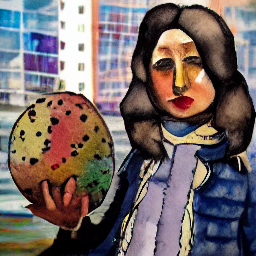}
\hspace{-1.2pt}\includegraphics[width=0.1\linewidth]{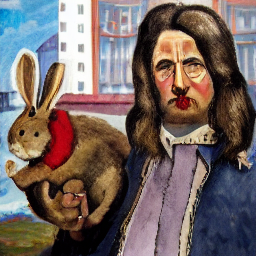}
\hspace{-1.2pt}\includegraphics[width=0.1\linewidth]{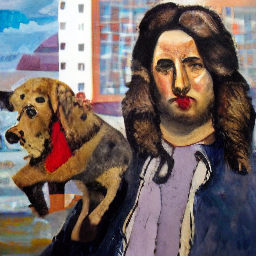}
\hspace{-1.2pt}\includegraphics[width=0.1\linewidth]{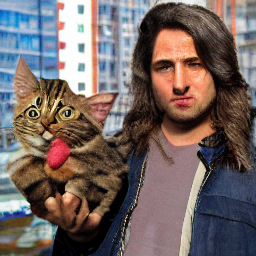}
\end{tabular}}
\vspace{1em}
\caption{Simultaneous editing of multiple attributes and objects of an image. Shown from left to right are  (i) input (ii) girl+watermelon (iii) woman+corgi  (iv) paint+cat+old woman  (v) paint + boy+ big egg     (vi) paint + man + rabbit  (vii) paint + man + dog  (viii)  man+cat\label{fig:simultaneous}}
\end{figure}
\section{Text Driven Manipulation with LDEdit}
In this section, we show how LDMs trained for text-to-image generation can be adapted for image manipulation.  Our main idea is to use a common  shared latent representation between the source image and the desired target, which is made possible by a deterministic diffusion process.     
The source image $\small{x_{\mbox{src}} }$ is mapped to a latent code $z_0$ by  
the encoder $\mathcal{E}$, and 
forward diffusion is performed until the time step $t_{stop}<T$ using DDIM sampling, conditioned on the source text prompt $y_{src}$ as:
\small
\begin{gather*}
\label{eq:ldedit_forward}
z_{t+1} = \sqrt{\alpha_{t+1}}\left(\frac{z_{t} - \sqrt{1-\alpha_t}\epsilonb_{\theta}(z_t,t,\tau_{\tilde{\theta}}(y_{src}))}{\sqrt{\alpha_{t}}}\right) +  \sqrt{1 - \alpha_{t+1}}\bm{\epsilon}_{{\theta}}(z_{t}, t,\tau_{\tilde{\theta}}(y_{src}))) 
\end{gather*}
\normalsize
The reverse diffusion conditioned on the target text prompt $y_{tar}$  starts from the same noised latent code $z_{t_{stop}}$ to arrive at $\hat{z}_0$:
\small
\begin{gather}
\label{eq:ldedit_reverse}
z_{t-1} = \sqrt{\alpha_{t-1}}\left(\frac{z_{t} - \sqrt{1-\alpha_t}\epsilonb_{\theta}(z_t,t,\tau_{\tilde{\theta}}(y_{tar}))}{\sqrt{\alpha_{t}}}\right) +  \sqrt{1 - \alpha_{t-1}}\bm{\epsilon}_{\theta}(y_{t}, t,\tau_{\tilde{\theta}}(y_{tar})) 
\end{gather}
\normalsize
Due to deterministic sampling, a near cycle-consistency is automatically maintained between source and target images~\cite{su2022dual}.   
Fig.~\ref{fig:method}~a) provides an overview of our approach, with an example where a source image with $y_{src}$ 'a yellow bus', is transformed according to the $y_{tar}$ 'a red bus' in a straightforward way. The visualized results obtained by decoding latents sampled in $[1,t_{stop}]$ during the forward and reverse diffusion process demonstrate the gradual transformation in the reverse process. Additionally, we can also introduce controlled stochasticity by varying $\small{\eta}$ \eqref{eq:eta}, which can produce diverse outputs as seen in Fig.~\ref{fig:method} c), with magnitude of $\small{\eta}$ controlling consistency with the original image. Further, Fig.~\ref{fig:method}~b) shows that changing the number of DDIM steps can also lead to some variance in our results. In the following section,  we demonstrate that this technique can accomplish a variety of image manipulation tasks using the pretrained LDM, in a zero-shot fashion without further optimization or fine-tuning.  
\section{Experiments}
We perform all our experiments with different image manipulation tasks using the text-to-image LDM with a downsampling factor of 8 pretrained using the openly available  LAION dataset~\cite{schuhmann2021laion} containing open-domain image-text pairs. We do not  fine-tune this model for any task. We set $t_{stop}\in[300, 640]$ out of the total 1000 steps and use fewer (20-80) steps between $[1, t_{stop}]$  in the deterministic forward and reverse diffusion. 
We perform experiments on both class-specific and open-domain images and compare with VQ+CLIP~\cite{crowson2022vqgan} which is versatile to handle general manipulation tasks. In addition, we also compare with class-specific approaches~\cite{patashnik2021styleclip,xia2021tedigan} and fine-tuned models~\cite{gal2021stylenada,kim2021diffusionclip} on the domain-specific tasks. Comparisons with the baseline-methods and run-time comparisons are performed with images of dimension $256\times256$ %

We first demonstrate our method on the task of manipulating an image of a yellow bus  according to the target prompts: `a tram', `a truck' and `a red steam engine'. Fig.~\ref{fig:bus} illustrates the results of this manipulation. The results indicate that LDEdit is able to manipulate the input according to the target texts even with a simple DDIM forward and reverse process with $\small{\eta=0}$. Further, by increasing $\small{\eta}$, our method is able to generate an assortment of diverse samples that are consistent with the pose of the yellow bus in the input image. The diversity increases as the parameter $\small{\eta}$ is increased. We also  illustrate the results obtained by  VQGAN+CLIP~\cite{crowson2022vqgan} on this task using two sets of hyper-parameters for comparison. While
\cite{crowson2022vqgan} can successfully transform the input image to that of `a tram', we were unable to obtain satisfactory results for the other two tasks, despite manual hyper-parameter tuning.

We further test our approach on manipulating images from diverse classes using test images from \cite{kim2021diffusionclip}.  We compare our performance with the generic approach of VQGAN+CLIP~\cite{crowson2022vqgan} and DiffusionCLIP~\cite{kim2021diffusionclip}, a state of the art method using class-specific models fine-tuned for the specific target texts.  Fig. \ref{fig:dclip} illustrates the results of this experiment. As DiffusionCLIP uses specific fine-tuned models on these tasks, it can effortlessly accomplish the desired manipulations. On the other hand, VQGAN+CLIP struggles to achieve desired changes when the target is highly different from the input.  Despite not being fine-tuned for the specific tasks, our LDEdit can accomplish the manipulations quite well. The task of manipulating a stroke image according to the target prompts is particularly challenging, as the input image lacks details. Handling such manipulation requires introducing stochasticity in the forward process, without which it is not possible to produce the desired edits.  

We further perform multiple manipulation tasks on face images, including semantic (multi)-attribute manipulation, style transfer, domain manipulation and compare with the recent state-of-the-art methods which are trained for face manipulation \cite{kim2021diffusionclip,patashnik2021styleclip,gal2021stylenada,xia2021tedigan}. The StyleGAN based methods~\cite{patashnik2021styleclip,gal2021stylenada,xia2021tedigan} employ the same encoders for GAN inversion as per the original setting in their work. Further, we include comparison with CLIP-Styler~\cite{kwon2021clipstyler} a CLIP guided texture manipulation approach, and VQGAN+CLIP~\cite{crowson2022vqgan} which can perform flexible image manipulation.  Fig.~\ref{fig:faces} illustrates our results.
While StyleGAN inversion based approaches \cite{patashnik2021styleclip,gal2021stylenada,xia2021tedigan} can manipulate semantic attributes see Fig.\ref{fig:faces}~c), they  struggle to reconstruct face images in atypical poses, see Fig.\ref{fig:faces}~a). Unexpected details present in the original image such as hand on the face are completely removed or distorted in the reconstructions. Since such atypical faces are hardly encountered  during training, StyleGAN inversion results in a high representation error. Similarly, it is hard to transfer to a different style \eg a watercolour painting, or domain \eg zombie using StyleGAN latent space search alone Fig.\ref{fig:faces}~a) and b). StyleGAN-NADA instead enable  these manipulations using domain-specific fine-tuning. On the other hand, ClipStyler~\cite{kwon2021clipstyler} can only accomplish global texture manipulations, and the result may drift away from the original colour palette. Among the compared methods, 
LDEdit, DiffusionCLIP~\cite{kim2021diffusionclip} and VQGAN+CLIP\cite{crowson2022vqgan}  accomplish the different manipulation tasks in addition to achieving good reconstructions, preserving identity better than GAN inversion based methods.  Interestingly, though VQGAN+CLIP and LDEdit are trained on generic images, these methods  are still able to perform on par with state of the art fine-tuned DiffusionCLIP~\cite{kim2021diffusionclip} on these tasks.

\begin{figure}[h]
\small
\resizebox{0.99\linewidth}{!}{\begin{tabular}{l}
\quad Input: girl+ball\qquad\quad girl+dog\qquad\qquad teen-girl+dog\qquad\quad woman+dog\qquad\quad old woman+dog
\\
\includegraphics[width=0.2\linewidth]{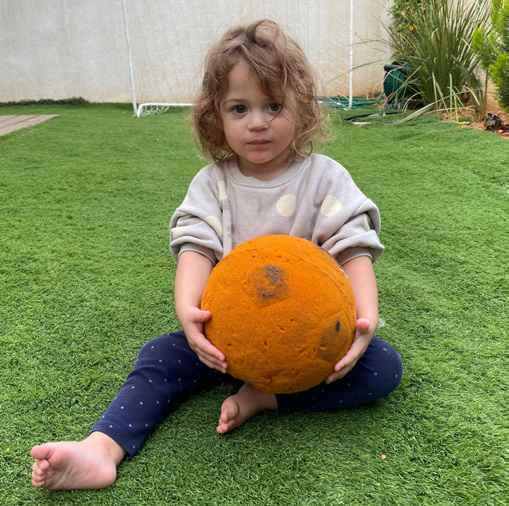}
\includegraphics[width=0.2\linewidth]{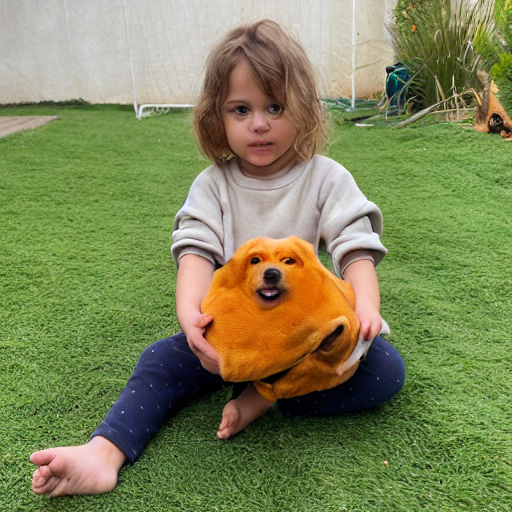}
\includegraphics[width=0.2\linewidth]{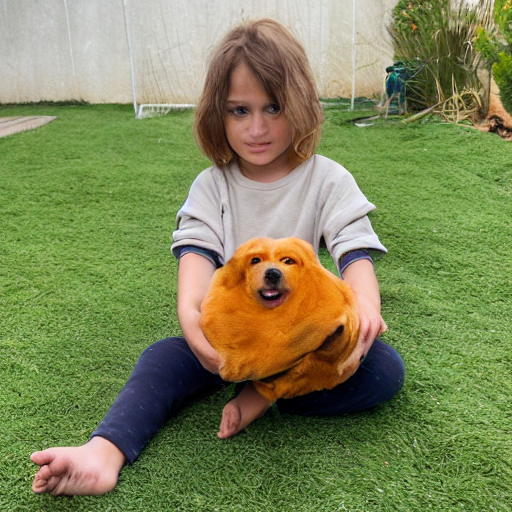}
\includegraphics[width=0.2\linewidth]{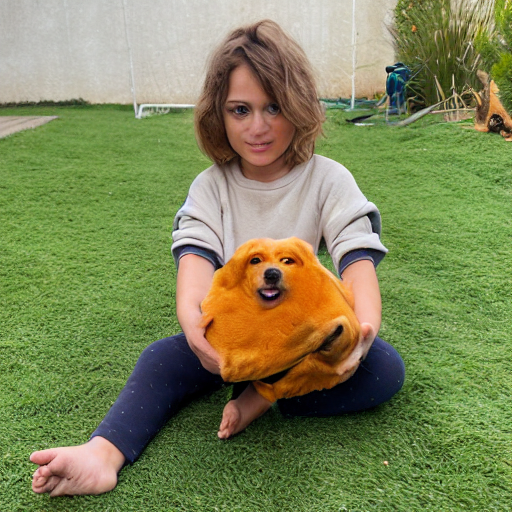}
\includegraphics[width=0.2\linewidth]{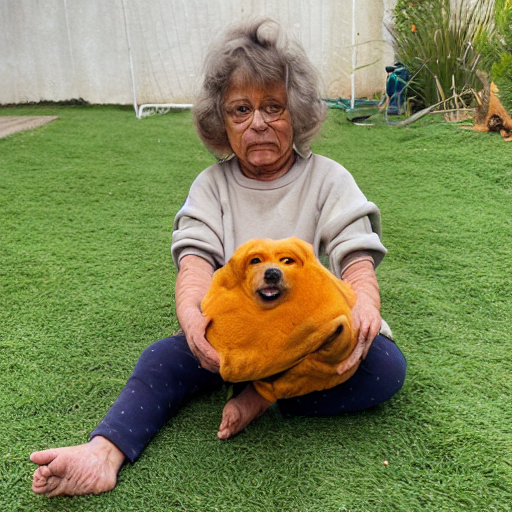}
\\
\quad Input: girl+ball\qquad boy+basketball\qquad teen-boy+basketball\quad man+basketball\quad old man+basketball
\\
\includegraphics[width=0.2\linewidth]{images/supple_child_ball/ground_truth.png}
\includegraphics[width=0.2\linewidth]{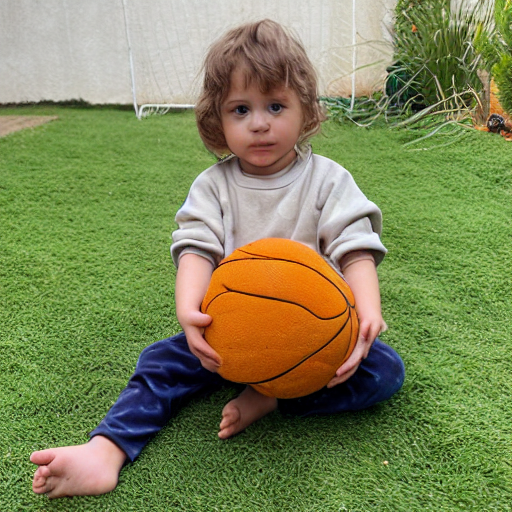}
\includegraphics[width=0.2\linewidth]{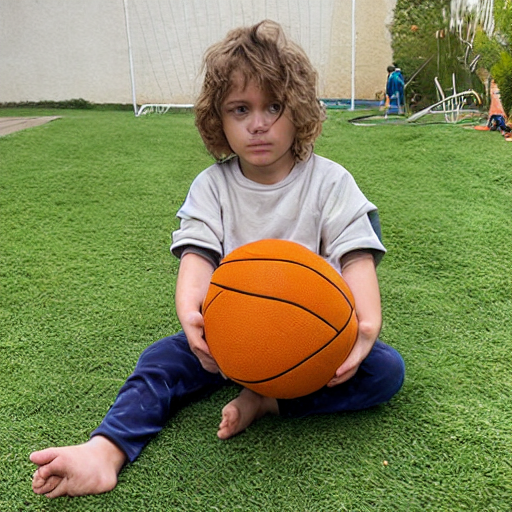}
\includegraphics[width=0.2\linewidth]{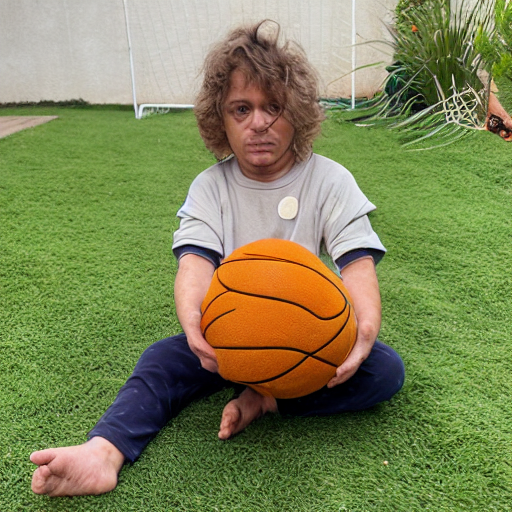}
\includegraphics[width=0.2\linewidth]{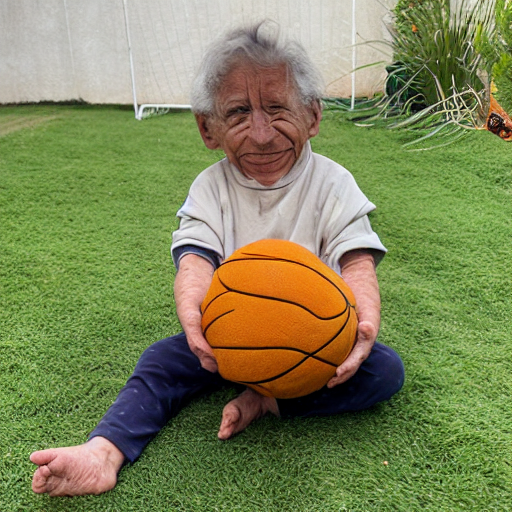}
\\
Input: 2 girls+hats \qquad Picasso style\qquad 2 women+hats\qquad 2 old women+hats\quad~Photo+2 babies+hats
\\
\includegraphics[width=0.2\linewidth]{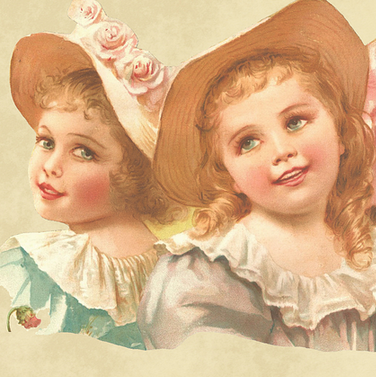}
\includegraphics[width=0.2\linewidth]{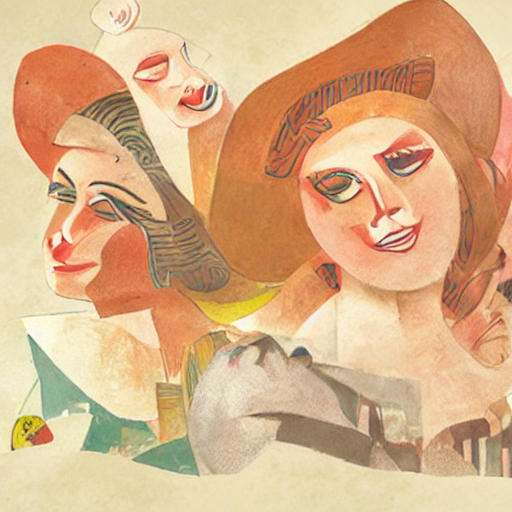}
\includegraphics[width=0.2\linewidth]{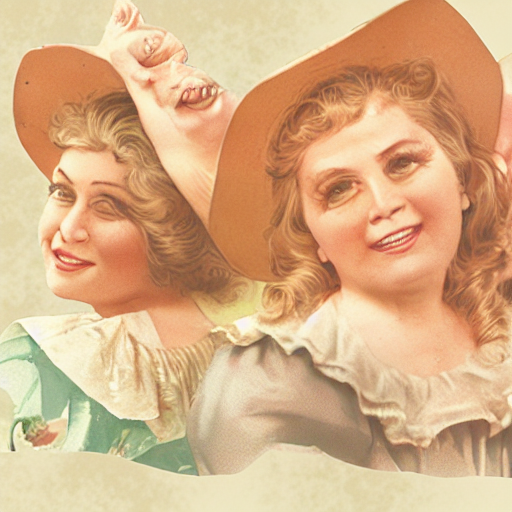}
\includegraphics[width=0.2\linewidth]{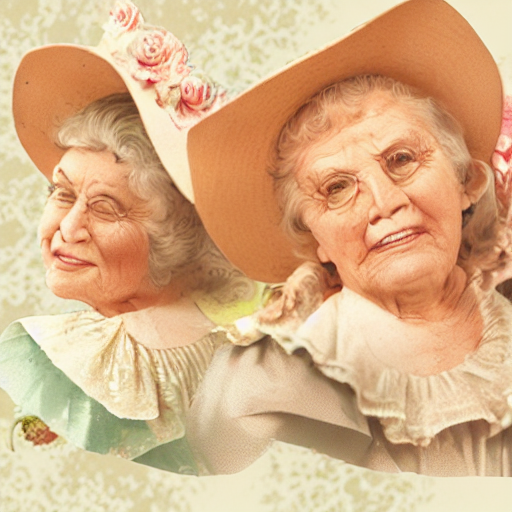}
\includegraphics[width=0.2\linewidth]{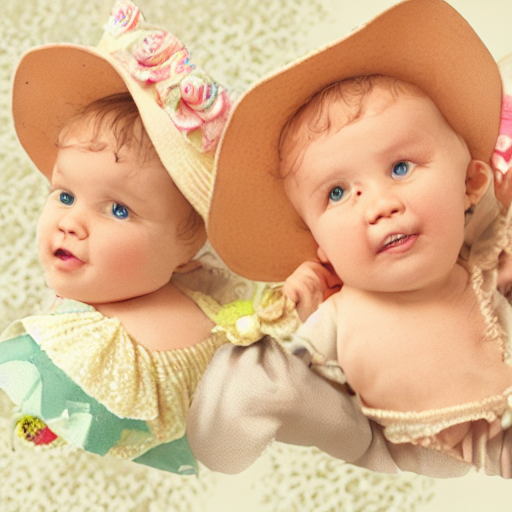}\\
\quad Input: a horse \qquad\qquad a zebra\qquad\qquad~~~a donkey\qquad \qquad~~~~~a bear\qquad \qquad\qquad~~~~a wolf
\\
\includegraphics[width=0.2\linewidth]{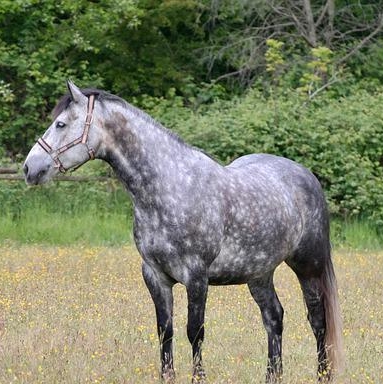}
\includegraphics[width=0.2\linewidth]{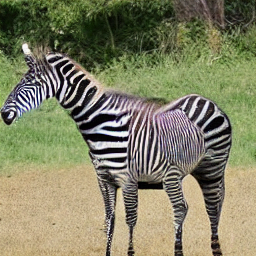}
\includegraphics[width=0.2\linewidth]{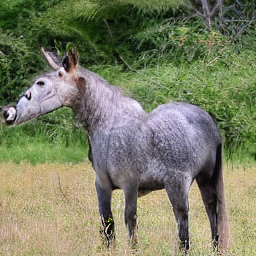}
\includegraphics[width=0.2\linewidth]{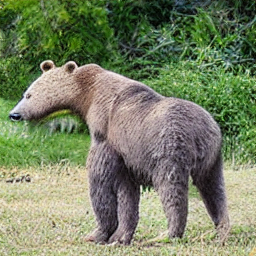}
\includegraphics[width=0.2\linewidth]{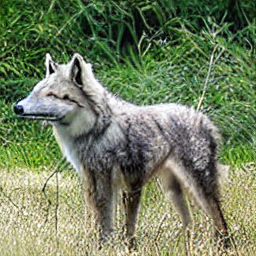}
\end{tabular}}
\vspace{1em}
\caption{Results of image manipulation using LDEdit \label{fig:girl_ball}}
\end{figure}
\begin{figure}[htb]
\small
\resizebox{0.99\linewidth}{!}{\begin{tabular}{l}
\quad Input: a deer\qquad Target: a deer with antlers\qquad\quad Input: a dog\qquad\quad Target: a lion\\
\includegraphics[width=0.16\linewidth]{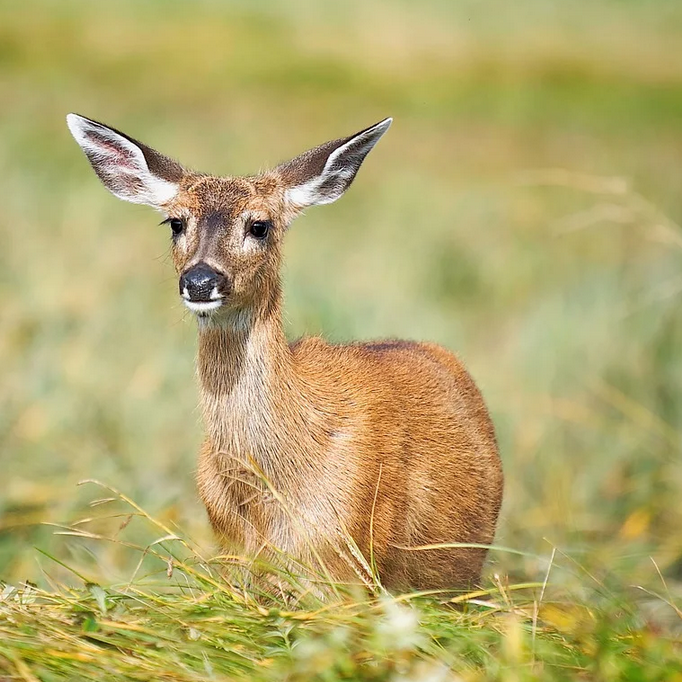}
\includegraphics[width=0.16\linewidth]{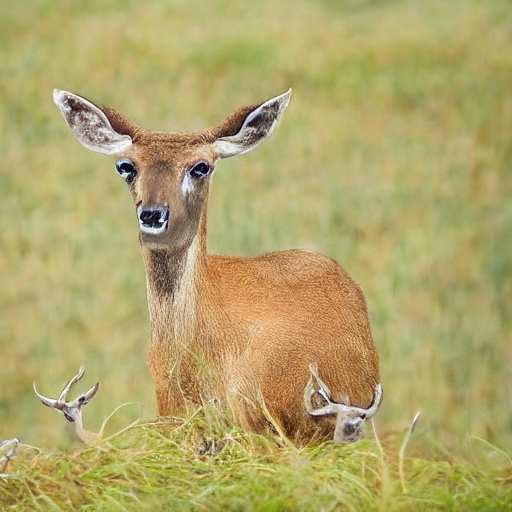}
\includegraphics[width=0.16\linewidth]{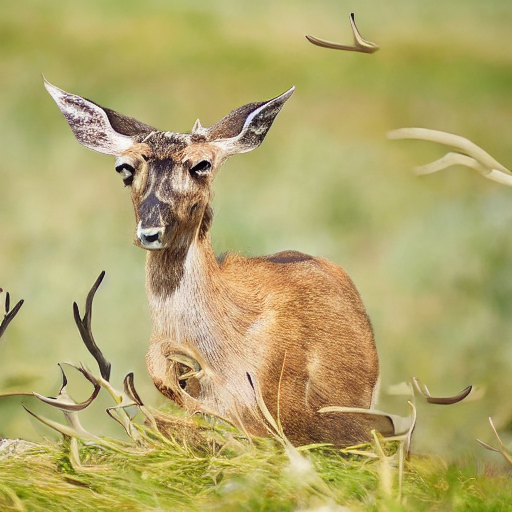}~
\includegraphics[width=0.16\linewidth]{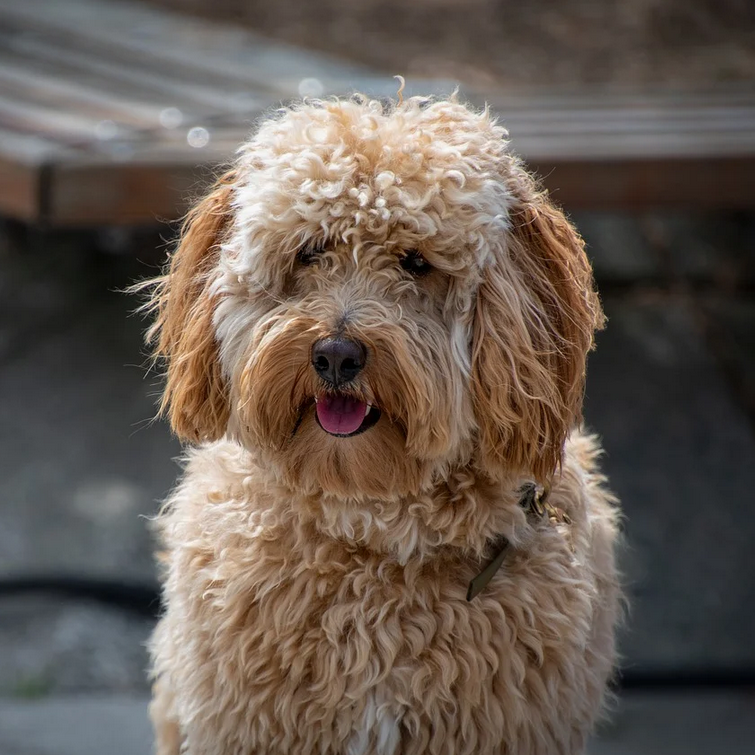}
\includegraphics[width=0.16\linewidth]{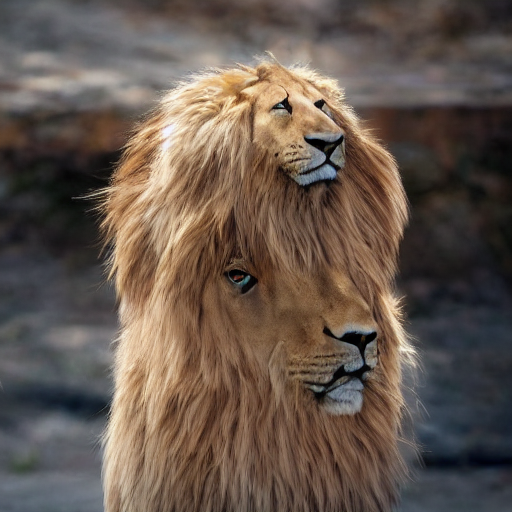}
\includegraphics[width=0.16\linewidth]{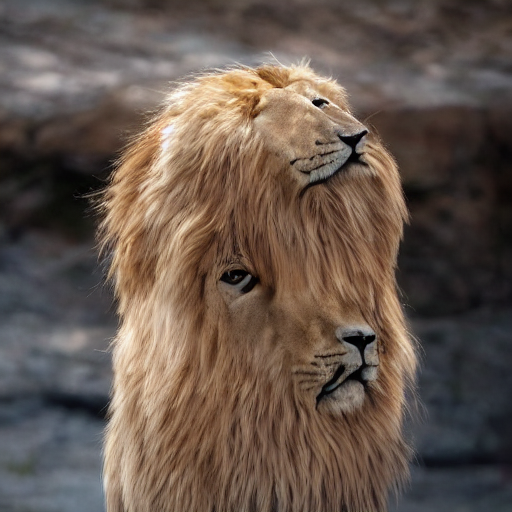}
\\
Input:girl+dog\qquad\qquad Target: girl+baby\qquad\qquad Input: girl+dog\qquad\quad Target: girl+cat\\
\includegraphics[width=0.16\linewidth]{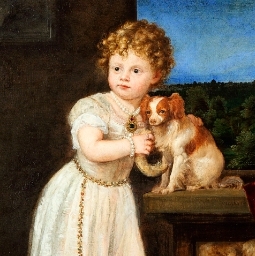}
\includegraphics[width=0.16\linewidth]{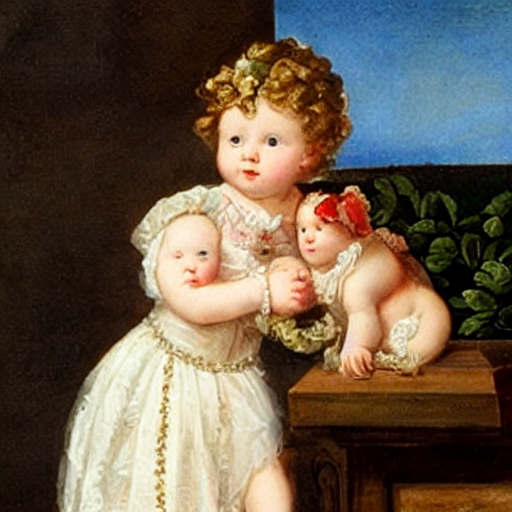}
\includegraphics[width=0.16\linewidth]{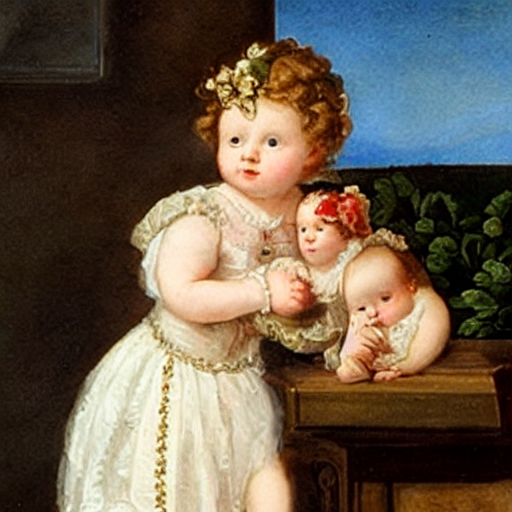}~
\includegraphics[width=0.16\linewidth]{masked_edit_figs/glide_painting_girl_with_dog.jpg}
\includegraphics[width=0.16\linewidth]{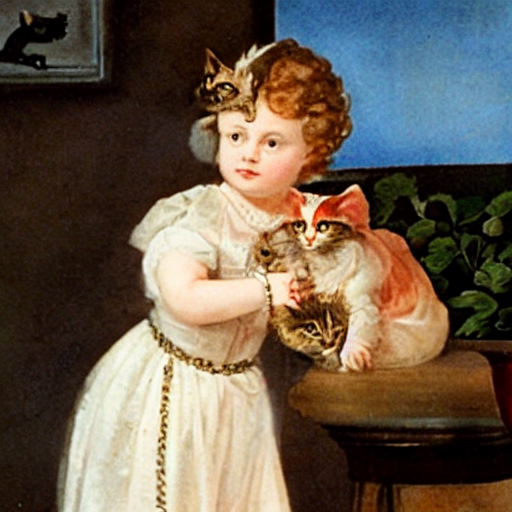}
\includegraphics[width=0.16\linewidth]{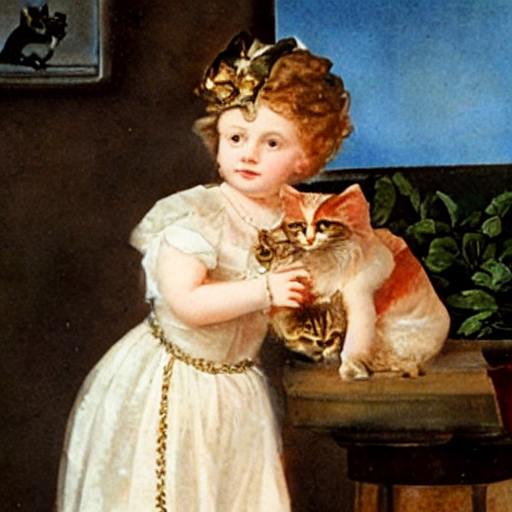}
\\
\end{tabular}}
\vspace{1em}
\caption{Failure cases of image manipulation using LDEdit \label{fig:failure}}
\end{figure}
\begin{figure}[h]
\small
\resizebox{0.99\linewidth}{!}{\begin{tabular}{l}
Input: girl+dog\qquad\enskip Mask\phantom{12345678} girl+cat\phantom{1234567}girl+flowers\phantom{1234}girl+monkey\phantom{12345}girl+baby\\
\includegraphics[width=0.16\linewidth]{masked_edit_figs/glide_painting_girl_with_dog.jpg}
\includegraphics[width=0.16\linewidth]{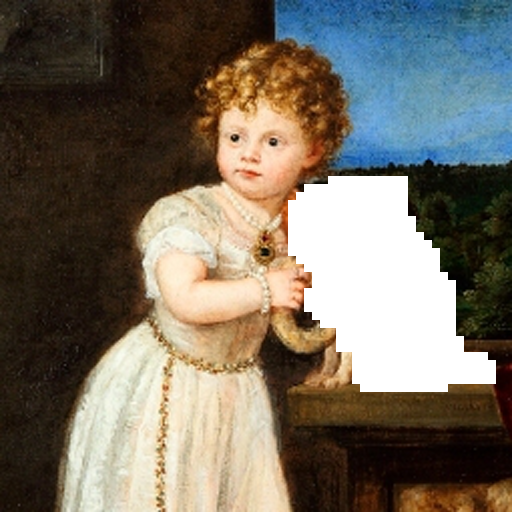}
\includegraphics[width=0.16\linewidth]{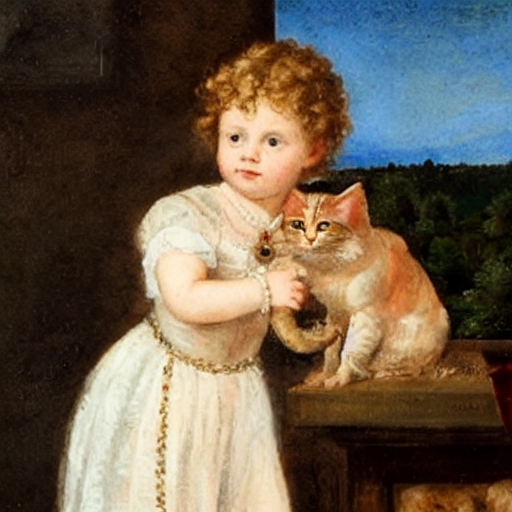}
\includegraphics[width=0.16\linewidth]{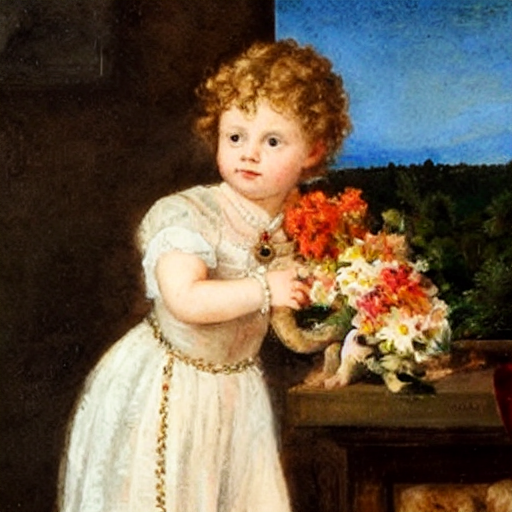}
\includegraphics[width=0.16\linewidth]{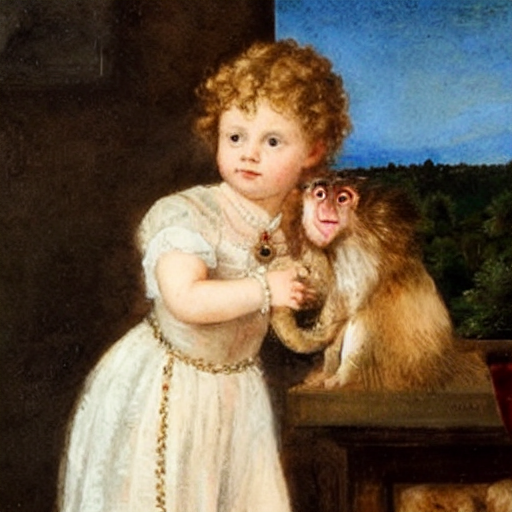}
\includegraphics[width=0.16\linewidth]{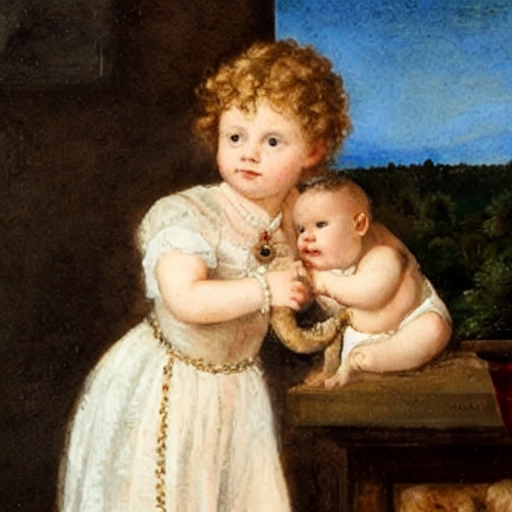}
\\
\end{tabular}}
\vspace{1em}
\caption{Masked image manipulation using LDEdit \label{fig:medit}}
\end{figure}

It is also possible to achieve further challenging manipulations involving simultaneous changes in  multiple attributes, local manipulations and artistic style changes  as seen in Fig. \ref{fig:simultaneous}. 
While the LDM model is trained on generation of images of dimension $256{\times}256$, due to fully convolutional nature of the autoencoder, our method can be applied on images of higher resolution using the same model. 
Fig.~\ref{fig:girl_ball} shows further example results of image manipulation using LDEdit, with image resolution $512\times512$. It is seen that our method can achieve varied transformations in a straightforward way. The first two rows show simultaneous manipulation of the girl and the ball. The third row shows style transfer to a painting or a photo and semantic manipulating the age of two girls. Interestingly, LDEdit can effect such transformations with a little or no stochasticity, such that the background remains largely unaffected. The final row shows manipulating a horse to other species, \eg
 a zebra, a donkey, a bear, and a wolf. These transformations required a higher $\eta$ of 0.3 for zebra and donkey, and $\eta$ of 0.8 for bear and wolf. However, higher values of $\eta$ result in more changes in the background.
 \begin{figure}[t]
\scriptsize
Input stroke\qquad old man\qquad woman\qquad pixar woman\quad van Gogh\qquad\enskip old man\qquad woman\qquad pixar woman\quad van Gogh\\
\includegraphics[width=0.11\linewidth]{images/inputs/stroke1.png}
\includegraphics[width=0.11\linewidth]{images/teaser/stroke1_oldman_ddpm_1.png}\hspace{-1.2pt}
\includegraphics[width=0.11\linewidth]{images/teaser/stroke1_woman_ddpm_11.png}\hspace{-1.2pt}
\includegraphics[width=0.11\linewidth]{images/ours/ddpm3_stroke_Pixar_woman_639_804.png}\hspace{-1.2pt}
\includegraphics[width=0.11\linewidth]{images/ours/gogh_ddpm1_stroke_woman_639_804.png}
\includegraphics[width=0.11\linewidth]{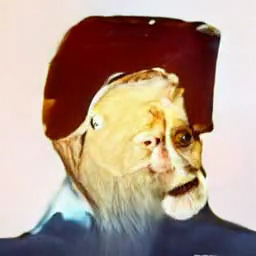}\hspace{-1.2pt}
\includegraphics[width=0.11\linewidth]{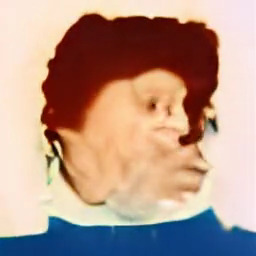}\hspace{-1.2pt}
\includegraphics[width=0.11\linewidth]{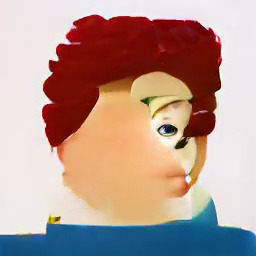}\hspace{-1.2pt}
\includegraphics[width=0.11\linewidth]{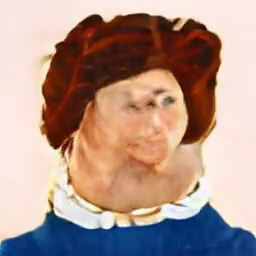}\\
\phantom{123}Input\qquad\quad red brick\qquad wooden\qquad Asian temple\qquad+snow\qquad\quad red brick\qquad wooden\qquad Asian temple\qquad+snow\\
\includegraphics[width=0.11\linewidth]{images/inputs/church3.png}
\includegraphics[width=0.11\linewidth]{images/teaser/temple3/redbrick.png}\hspace{-1.2pt}
\includegraphics[width=0.11\linewidth]{images/teaser/temple3/woodenhouse.png}\hspace{-1.2pt}
\includegraphics[width=0.11\linewidth]{images/teaser/temple3/asiantemple.png}\hspace{-1.2pt}
\includegraphics[width=0.11\linewidth]{images/teaser/temple3/res_snow_asian.png}
\includegraphics[width=0.11\linewidth]{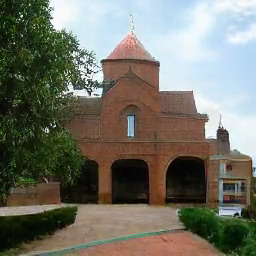}\hspace{-1.2pt}
\includegraphics[width=0.11\linewidth]{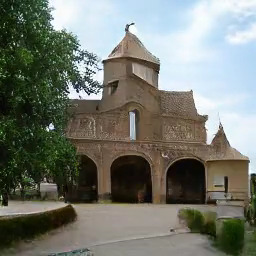}\hspace{-1.2pt}
\includegraphics[width=0.11\linewidth]{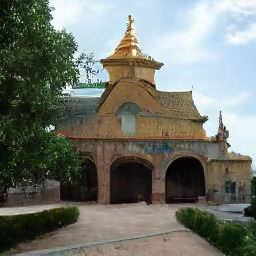}\hspace{-1.2pt}
\includegraphics[width=0.11\linewidth]{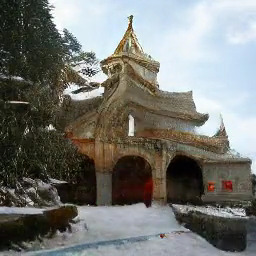}\\
 \phantom{123}Input\qquad\quad an eagle\qquad a kingfisher\qquad crow+tree\quad crow+sketch\quad an eagle\qquad a kingfisher\qquad crow+tree\quad crow+sketch\\
\includegraphics[width=0.11\linewidth]{images/teaser/pigeon/pigeon.png}
\includegraphics[width=0.11\linewidth]{images/teaser/pigeon/eagle.png}\hspace{-1.2pt}
\includegraphics[width=0.11\linewidth]{images/teaser/pigeon/kingfisher.png}\hspace{-1.2pt}
\includegraphics[width=0.11\linewidth]{images/teaser/pigeon/crow_on_tree.png}\hspace{-1.2pt}
\includegraphics[width=0.11\linewidth]{images/teaser/pigeon/crow_illustration.png}
\includegraphics[width=0.11\linewidth]{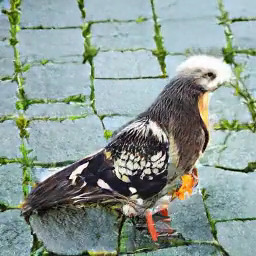}\hspace{-1.2pt}
\includegraphics[width=0.11\linewidth]{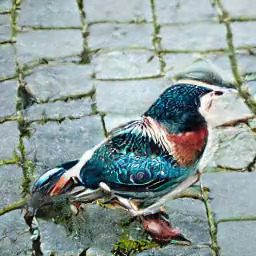}\hspace{-1.2pt}
\includegraphics[width=0.11\linewidth]{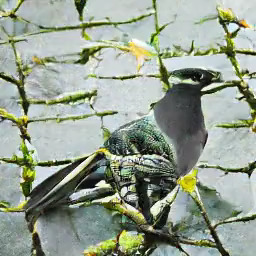}\hspace{-1.2pt}
\includegraphics[width=0.11\linewidth]{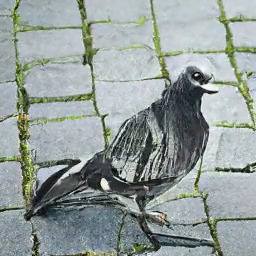}\\
 \phantom{123}Input\quad\enskip girl+w.melon\quad lady+corgi\enskip~~~ man+cat\enskip~~ old lady+cat+paint~ girl+w.melon\enskip lady+corgi\enskip~~ man+cat\enskip old lady+cat+paint\\
\includegraphics[width=0.11\linewidth]{images/pix2/pixabay2.png.png}
\includegraphics[width=0.11\linewidth]{images/pix2/woman_to_child_0.3_0.png}
\hspace{-1.2pt}\includegraphics[width=0.11\linewidth]{images/pix2/woman_to_woman_corgi_0.1_0.png}\hspace{-1.2pt}
\includegraphics[width=0.11\linewidth]{images/pix2/woman_to_man_cat_0.3_4.png}
\hspace{-1.2pt}\includegraphics[width=0.11\linewidth]{images/pix2/woman_to_old_woman_cat_0.3_2.png}
\includegraphics[width=0.11\linewidth]{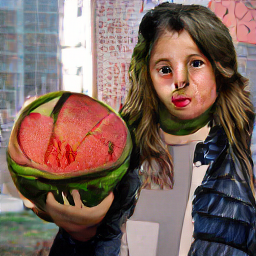}
\hspace{-1.2pt}\includegraphics[width=0.11\linewidth]{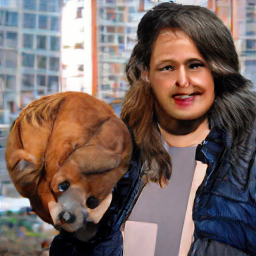}
\hspace{-1.2pt}\includegraphics[width=0.11\linewidth]{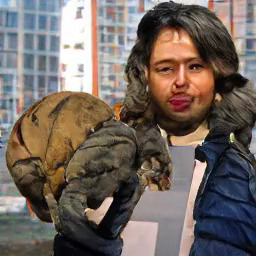}
\hspace{-1.2pt}\includegraphics[width=0.11\linewidth]{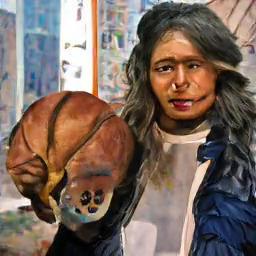}\\
\phantom{1234}Input\phantom{1234}$\xleftarrow{\hspace*{0.138\linewidth}}\text{ LDEdit (Ours) } \xrightarrow{\hspace*{0.138\linewidth}}$~$\xleftarrow{\hspace*{0.135\linewidth}} \text{VQGAN+CLIP~\cite{crowson2022vqgan}} \xrightarrow{\hspace*{0.135\linewidth}}$\\
\caption{Comparison of LDEdit with VQGAN+CLIP~\cite{crowson2022vqgan}. Best results out of 4 samples are shown for LDEdit and the best result out of 8 samples are shown for VQGAN+CLIP.\label{fig:vqclipsupple}}
\end{figure}\vspace{2pt}\\
 \noindent\textbf{Comparisons with VQGAN+CLIP~\cite{crowson2022vqgan}} We provide more visual comparisons with  VQGAN+CLIP for general text driven image manipulation in Fig.~\ref{fig:vqclipsupple}. While VQGAN+CLIP can successfully effect changes in the input image of a building as per the target text prompts, its performance suffers in more difficult manipulations such as translating from an input stroke, or performing simultaneous local manipulations.  In contrast, our LDEdit is able to perform these desired manipulations.\vspace{2pt}\\ \noindent\textbf{Failure Cases} In some cases, our method may fail to produce desired manipulations as seen in Fig.~\ref{fig:failure}. With an input text prompt of `a deer with antlers', we obtain manipulated images where the antlers are misplaced. In other cases, we obtain features of target objects additionally in undesired locations, such as a baby face on the girl's hand, or a cat face in the hair and in the background picture frame. These undesired effects can be avoided by using a mask, which can aid in localization of edits.\vspace{2pt}\\ \noindent\textbf{Editing with Masks}
Our method can be modified to include a user-specified mask which specifies the regions where significant changes are needed. Similar mask-guided editing has also been shown in  \cite{avrahami2021blended,nichol2021glide}. The  user-specified mask is also down-sampled such that it has the same spatial extent as the latent code. Let $z_{t_{stop}}$ be the latent code after forward diffusion, the desired localized edit can be obtained by performing the reverse diffusion process on multiple copies of $z_{t_{stop}}$,  by changing the target text for the respective masked regions. For seamless blending of the masked and unmasked regions, the latent code corresponding to the two regions are combined at each diffusion step. This even allows us to specify different levels of stochasticity for the  different regions. Fig.~\ref{fig:medit} shows the result of such mask masked editing. We can see that our approach successfully results in a seamless local editing, without requiring expensive optimization.\\ 
\begin{figure}[t]
\footnotesize
   \centering
    \includegraphics[width=\linewidth]{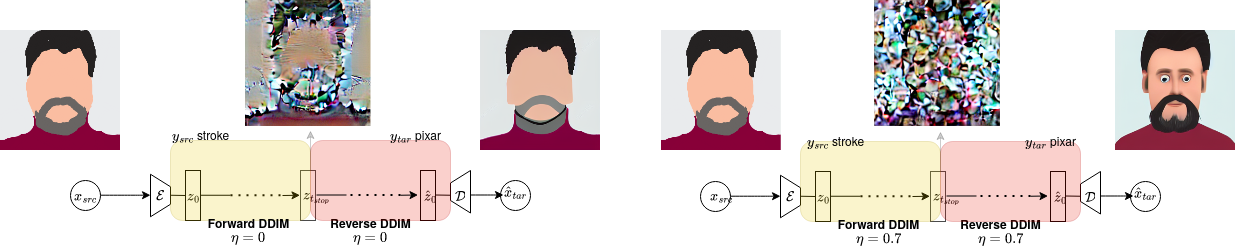}\\
    a)~Deterministic diffusion\hspace{100pt}b)~DDIM with $\eta$\vspace{2pt}\\
    \includegraphics[width=0.098\linewidth]{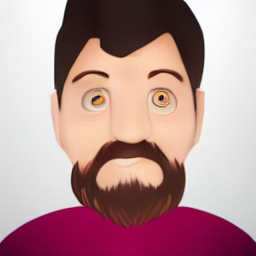}
   \hspace{-1.9pt}\includegraphics[width=0.098\linewidth]{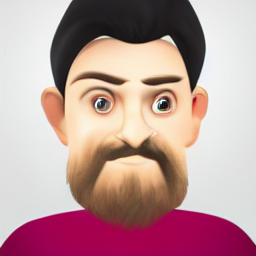}
      \hspace{-1.9pt}\includegraphics[width=0.098\linewidth]{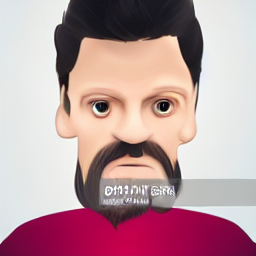}
      \hspace{-1.9pt}\includegraphics[width=0.098\linewidth]{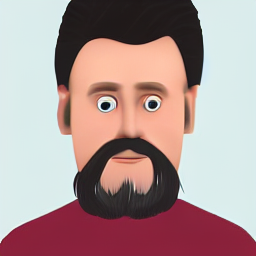}
        \hspace{-1.9pt}\includegraphics[width=0.098\linewidth]{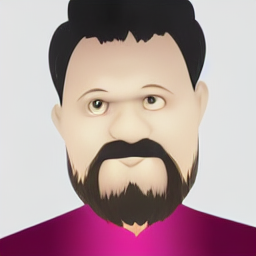}~
        \includegraphics[width=0.098\linewidth]{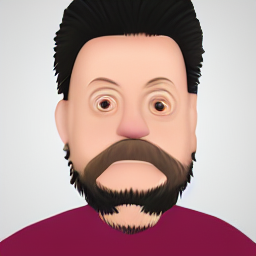}
   \hspace{-1.9pt}\includegraphics[width=0.098\linewidth]{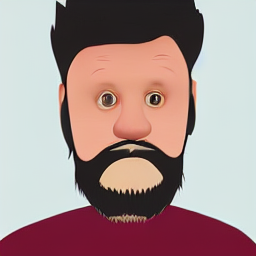}
      \hspace{-1.9pt}\includegraphics[width=0.098\linewidth]{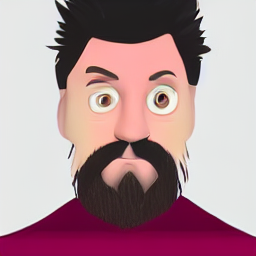}
      \hspace{-1.9pt}\includegraphics[width=0.098\linewidth]{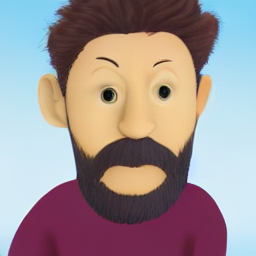}
        \hspace{-1.9pt}\includegraphics[width=0.098\linewidth]{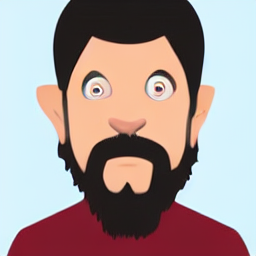}
        c)~Different samples $\eta=0.7$\hspace{100pt}d)~Different samples $\eta=0.9$\vspace{2pt}\\
    \caption{Effect of $\eta$ in diffusion process. Purely deterministic DDIM process cannot achieve desired target when the original input lacks details.\label{fig:eta}}
\end{figure}
\begin{figure}[htb]
\scriptsize
\ horse$\rightarrow$zebra\hspace{50pt}$\eta=0.2$\hspace{75pt}$\eta=0.3$\hspace{75pt}$\eta=0.6$\\\includegraphics[width=0.099\linewidth]{images/supple_horse/grey_horse3.jpg}
\includegraphics[width=0.099\linewidth]{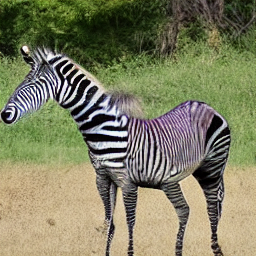}
\hspace{-1.5pt}\includegraphics[width=0.099\linewidth]{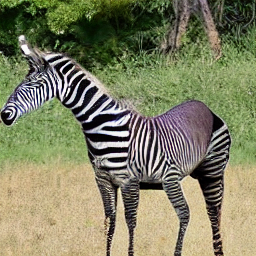}
\hspace{-1.5pt}\includegraphics[width=0.099\linewidth]{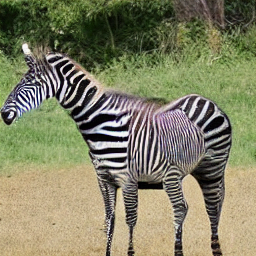}
\includegraphics[width=0.099\linewidth]{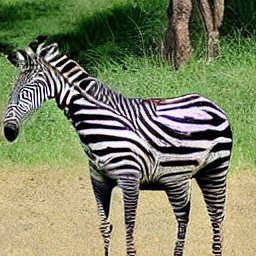}
\hspace{-1.5pt}\includegraphics[width=0.099\linewidth]{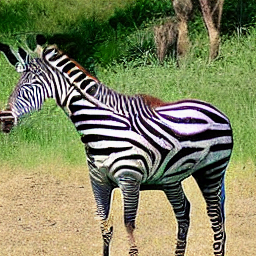}
\hspace{-1.5pt}\includegraphics[width=0.099\linewidth]{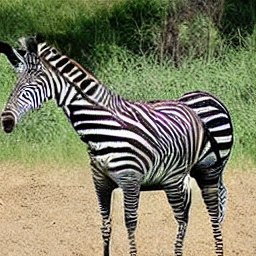}
\includegraphics[width=0.099\linewidth]{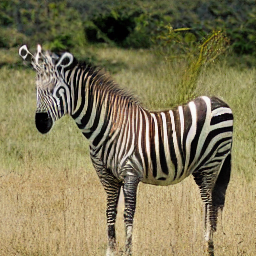}
\hspace{-1.5pt}\includegraphics[width=0.099\linewidth]{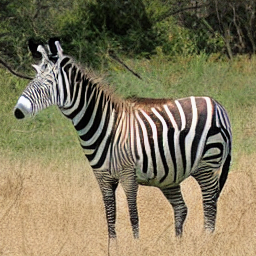}
\hspace{-1.5pt}\includegraphics[width=0.099\linewidth]{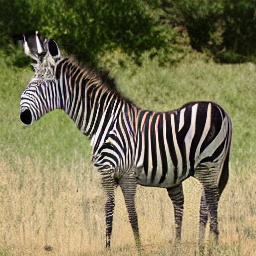}\\
 horse$\rightarrow$wolf\hspace{50pt}$\eta=0.3$\hspace{75pt}$\eta=0.5$\hspace{75pt}$\eta=0.8$\\\includegraphics[width=0.099\linewidth]{images/supple_horse/grey_horse3.jpg}
\includegraphics[width=0.099\linewidth]{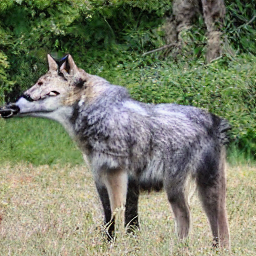}
\hspace{-1.5pt}\includegraphics[width=0.099\linewidth]{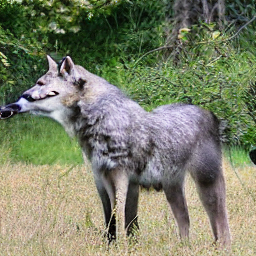}
\hspace{-1.5pt}\includegraphics[width=0.099\linewidth]{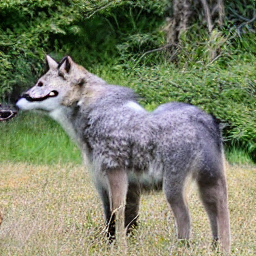}
\includegraphics[width=0.099\linewidth]{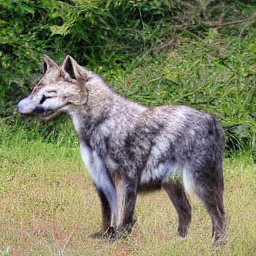}
\hspace{-1.5pt}\includegraphics[width=0.099\linewidth]{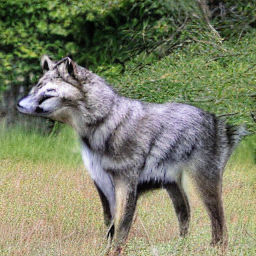}
\hspace{-1.5pt}\includegraphics[width=0.099\linewidth]{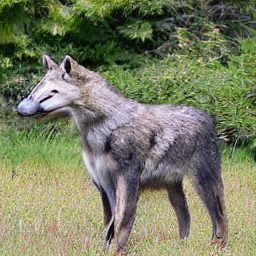}
\includegraphics[width=0.099\linewidth]{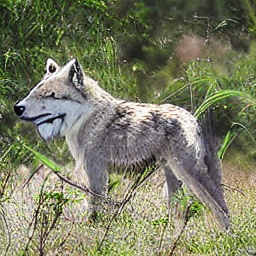}
\hspace{-1.5pt}\includegraphics[width=0.099\linewidth]{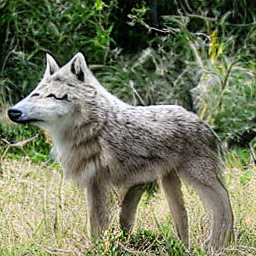}
\hspace{-1.5pt}\includegraphics[width=0.099\linewidth]{images/supple_horse/horse3_2wolf0.8_selected.png}\\
\hspace{10pt}dog$\rightarrow$fox\hspace{50pt}$\eta=0.1$\hspace{75pt}$\eta=0.4$\hspace{75pt}$\eta=0.8$\\
\includegraphics[width=0.099\linewidth]{images/inputs/dog1.png}
\includegraphics[width=0.099\linewidth]{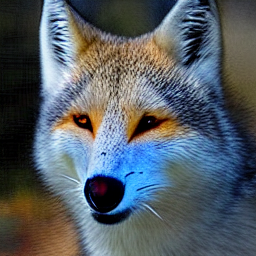}
\hspace{-1.5pt}\includegraphics[width=0.099\linewidth]{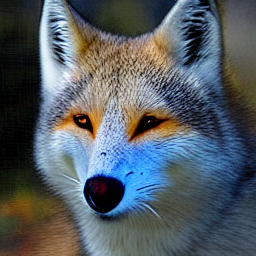}
\hspace{-1.5pt}\includegraphics[width=0.099\linewidth]{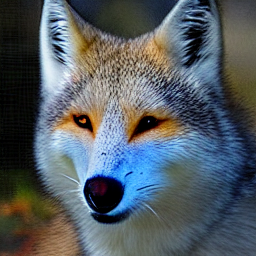}
\includegraphics[width=0.099\linewidth]{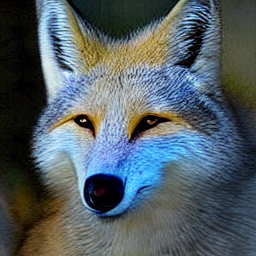}
\hspace{-1.5pt}\includegraphics[width=0.099\linewidth]{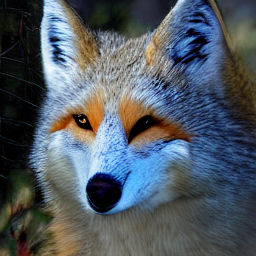}
\hspace{-1.5pt}\includegraphics[width=0.099\linewidth]{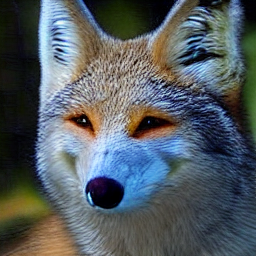}
\includegraphics[width=0.099\linewidth]{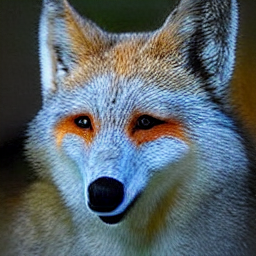}
\hspace{-1.5pt}\includegraphics[width=0.099\linewidth]{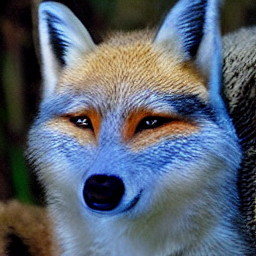}
\hspace{-1.5pt}\includegraphics[width=0.099\linewidth]{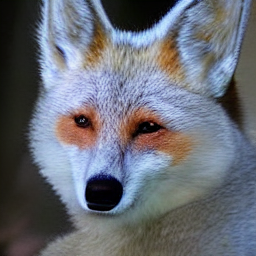}\\
\caption{Sample results for different $\eta$  using LDEdit. As the value of $\eta$ increases, the diversity of samples increases. \label{fig:multiple}}
\end{figure}\\
\noindent\textbf{Effect of Stochasticity}
In our approach, we proposed to perform a deterministic DDIM sampling, to ensure that a consistency is maintained with the original image. However, when the input image lacks details, such as a stroke image, doing a deterministic forward produces a latent code which lacks any details, see Fig.~\ref{fig:eta} a). On the other hand, introduction of stochasticity through $\eta$ can aid in hallucinating details not present in the original image, Fig.~\ref{fig:eta} b). With $\eta=1$, DDIM becomes equivalent to DDPM sampling, which results in more diverse samples. Note that our method may sometimes result in images with text like artifacts, as seen in Fig.~\ref{fig:eta} c). More example image manipulation of LDEdit by varying $\eta$ are shown in Fig.~\ref{fig:multiple}. As the value of $\eta$ increases, the diversity of samples improves. However, there are more perceptible changes in background, see rows 1 and 2 of Fig.~\ref{fig:multiple}.\vspace{2pt}\\ \noindent\textbf{User Study} 
We conduct  user studies to compare user preference of image manipulation results of  our method with VQGAN+CLIP~\cite{crowson2022vqgan} and DiffusionCLIP~\cite{kim2021diffusionclip}. Users participated in two surveys, where they were provided with source image, target text description and the results obtained with LDEdit and base-line method (VQGAN+CLIP or DiffusionCLIP) in a random order, and voted their preferred image manipulation using a survey platform.  We obtained a total of  1120 votes from 32 participants for comparing LDEdit with VQGAN+CLIP  and 950 votes from 38 participants  for comparing LDEdit with DiffusionCLIP. For comparison with both the baselines, we included a combination of face images  and general images (on manipulations demonstrated in DiffusionCLIP~\cite{kim2021diffusionclip} paper). On faces,  manipulated attributes include  makeup, tanned, curly hair, changing gender, domain change to zombie, neanderthal. We also include an example of  translating stroke image to pixar, neanderthal and van Gogh painting. On general image manipulation, we include manipulating an input building, bus, dog and a tennis ball. Additionally, for comparison with VQGAN+CLIP, we include examples of manipulating an image of a bird and multiple local object manipulations. In human evaluation, the results of LDEdit were preferred 83.87\% of the time in the survey comparing LDEdit with VQGAN+CLIP, whereas user preference  for LDEdit is 49.15\% in the survey comparing LDEdit with DiffusionCLIP.\vspace{2pt}\\
\textbf{Run-time~}Tab.~\ref{tab:timing} provides a comparison of GPU memory requirements and run-times of different text based image manipulation methods. The experiments were conducted on a computer with  AMD Ryzen 9 3950X 16-Core Processor and NVIDIA GeForce RTX 3090 with 24GB GPU memory. The run-times are highest for VQGAN+CLIP~\cite{crowson2022vqgan} (in the order of minutes), which requires an expensive optimization. Further, VQGAN+CLIP  requires different number of iterations to achieve the desired edit depending on the target prompt, leading to variable run-times.  The run-times of both DiffusionCLIP~\cite{kim2021diffusionclip} and our proposed LDEdit are significantly lower, with LDEdit having smaller run-times due to diffusion in smaller dimensional latent space.
It is to be noted that DiffusionCLIP~\cite{kim2021diffusionclip} needs to be fine-tuned for specific text prompts using a set of images ($\sim$ 30-50 images for each prompt), which takes $2-6$ minutes. Our method also scales well in terms of performing manipulations on multiple images in parallel, in contrast to VQGAN+CLIP, where manipulation on only 2 images could be performed in parallel.

\begin{table}[htb]
    \centering
    \resizebox{0.7\linewidth}{!}{
    \begin{tabular}{ccccc}
    \hline
         Method&\#images&GPU Memory&run-time&$(n_{for},n_{rev})$\\
         \hline
        LDEdit & 1 &8831MB&2.02s$\pm$ 5.58 ms&(25,25)\\
        LDEdit & 24 &16947MB&22.6$\pm$169ms&(25,25)\\
        LDEdit & 1 &8831MB&6.05s$\pm$35.6 ms&(75,75)\\
        LDEdit & 24 &16947MB&67.2s$\pm$704ms&(75,75)\\
        VQGAN+CLIP~\cite{crowson2022vqgan}&1&10413 MB &4-6 mins&--\\
        VQGAN+CLIP~\cite{crowson2022vqgan}&2&18933 MB &5-8 minutes&--\\
        DiffusionCLIP~\cite{kim2021diffusionclip}&1&5385MB&11.54s$\pm$66.3ms&(200,40)\\
        DiffusionCLIP~\cite{kim2021diffusionclip}&1&5385MB&4.01s$\pm$10.5ms&(40,40)\\
         DiffusionCLIP~\cite{kim2021diffusionclip}&24&15257MB&156.94s$\pm$470ms&(200,40)\\
        \hline\vspace{1.5pt}
    \end{tabular}}
    \caption{Comparing inference times and GPU memory usage of LDEdit with VQGAN+CLIP~\cite{crowson2022vqgan} and DiffusionCLIP~\cite{kim2021diffusionclip}. Images are of dimension $256\times 256$. $n_{for}$ and $n_{rev}$ refer to the number of forward and reverse diffusion steps. Mean and standard deviation of run-times over 10 runs are reported for LDEdit and DiffusionCLIP. }
    \label{tab:timing}
\end{table}
\section{Discussion and Conclusions}
We proposed LDEdit, a fast and flexible approach to open domain image manipulation using arbitrary text prompts. Our approach utilizes recent text-to-image latent diffusion model to achieve  zero-shot manipulation.
Experiments demonstrate that the proposed method can accomplish fast and diverse manipulation making  our approach a versatile tool to facilitate efficient user-guided editing. 
As with other image generation and manipulation methods, there is a potential for LDEdit being misused by bad actors for generating deepfakes and doctored pictures for propaganda. Further, since LDEdit leverages a pretrained text to image latent diffusion model, our approach inherits the  inherent biases of its training  dataset, including, but not limited to gender, age, and ethnicity of people and cultural biases. 
a
\end{document}